\documentclass[11pt]{article}

\usepackage[preprint]{acl}

\usepackage{times}
\usepackage{latexsym}

\usepackage[T1]{fontenc}

\usepackage[utf8]{inputenc}

\usepackage{microtype}

\usepackage{inconsolata}

\usepackage{graphicx}

\usepackage{amsmath}
\usepackage{amssymb}
\usepackage{booktabs}
\usepackage{multirow}
\usepackage{algorithm}
\usepackage{algpseudocode}
\usepackage{enumitem}
\usepackage{xspace}
\usepackage{adjustbox}
\usepackage{makecell}
\usepackage{rotating}

\usepackage{tabularx}
\usepackage{array}
\usepackage{seqsplit}
\usepackage{fdsymbol}

\newcolumntype{P}[1]{>{\raggedright\arraybackslash}p{#1}}
\newcolumntype{Y}{>{\raggedright\arraybackslash}X}
\newcommand{\hp}[1]{\begingroup\ttfamily\seqsplit{#1}\endgroup}

\title{EvoFlux: Inference-Time Evolution of Executable Tool Workflows for Compact Agents}

\author{
 \textbf{Kushal Raj Bhandari$^{1}$}\footnotemark[1],
 \textbf{Ling Yue$^{1}$},
 \textbf{Ching-Yun Ko$^{2}$},
 \textbf{Dhaval Patel$^{2}$}, \\
 \textbf{Shaowu Pan$^{1}$},
 \textbf{Pin-Yu Chen$^{2}$}\thanks{Corresponding authors: \href{mailto:bhandk@rpi.edu}{bhandk@rpi.edu}, \href{mailto:pin-yu.chen@ibm.com}{pin-yu.chen@ibm.com}, \href{mailto:gaoj8@rpi.edu}{gaoj8@rpi.edu}},
 \textbf{Jianxi Gao$^{1}$}\footnotemark[1]
\\
 \textsuperscript{1}~Rensselaer Polytechnic Institute, Troy, NY 12180 USA\\
 \textsuperscript{2}~IBM Research, Yorktown Heights, NY 10598 USA
}

\begin{document}
\maketitle
\begin{abstract}
    Small language models (1.5B--4B parameters) offer low-cost, low-latency tool use but often produce plausible plans that fail during execution due to unavailable tools, missing arguments, or broken dependencies. We introduce \textsc{EvoFlux}, an inference-time search method that repairs tool plans through an evolutionary loop over structured graphs of tool calls and dependencies. EvoFlux executes candidate plans against live tool servers and uses execution errors to guide targeted edits, including replacing tools, correcting parameters, inserting missing steps, and removing redundant ones. On held-out MCP-Bench tasks spanning 28 live servers and 250 tools, EvoFlux improves execution success from roughly 3\% to 17--24\% across four small planners. SFT and SFT+DPO trained on the same 177 demonstrations and 624 preference pairs match or underperform the zero-shot baseline, while ReAct~\cite{yaoReActSynergizingReasoning2023} reaches higher peaks on stronger planners at 4--20$\times$ greater planner-token cost per solved task. Execution-guided inference-time repair reliably improves small tool-using agents when supervision is limited.
    Code is available at \url{https://github.com/IBM/Evoflux}.
\end{abstract}
\section{Introduction}
\vspace{-0.5em}
Language models are increasingly deployed as agents that call external tools rather than answer solely from memorized knowledge. Making a single tool call is now routine, but real tasks require chaining multiple calls into a \emph{workflow} where each step's output feeds the next. For example, answering ``find recent papers on retrieval-augmented generation and summarize the top three'' requires discovering a search tool, calling it with the right query, passing paper identifiers to a retrieval tool, and synthesizing a grounded answer. Each step must use an available tool, pass schema-conforming parameters, and consume outputs from earlier steps. Tool use thus becomes a \emph{workflow construction} problem \citep{yue2026static}, where the model must assemble a small program whose steps compile, execute, and carry evidence forward.

\begin{figure}[!t]
    \centering
    \includegraphics[width=\linewidth]{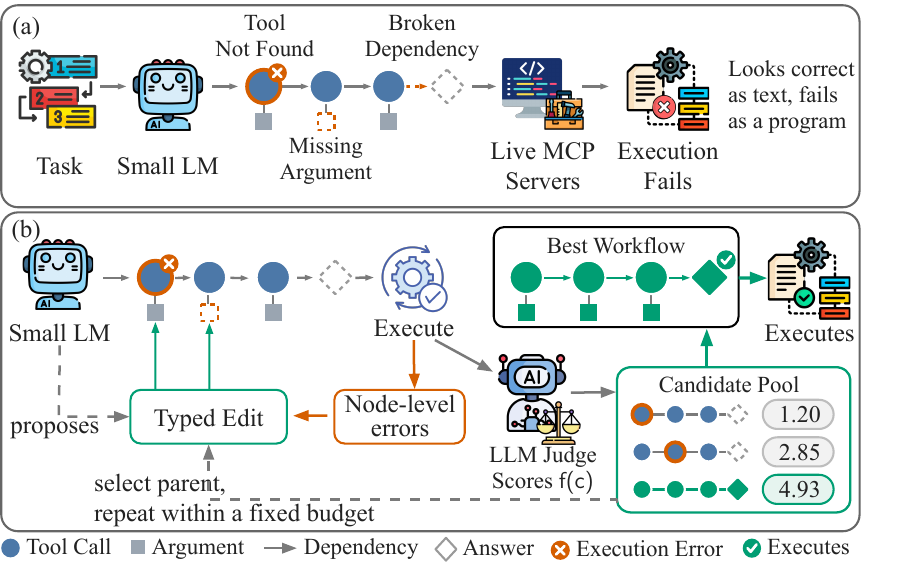}
    \caption{\textbf{(a)} Small LMs write tool workflows that look plausible but fail on live MCP servers.\textbf{(b)} \textsc{EvoFlux} executes each candidate, turns node-level errors into typed edits proposed by the same small model, and keeps a judge-scored candidate pool from which it selects parents and returns the best workflow, without any weight updates.}
    \label{fig:motivation}
    \vspace{-2.0em}
\end{figure}

The Model Context Protocol (MCP) standardizes tool discovery and invocation across servers. Tool-use benchmarks have accordingly moved from isolated function-calling in ToolBench \citep{qinToolLLMFacilitatingLarge2024} and Gorilla \citep{gonzalezGorillaLargeLanguage2024} toward multi-step tasks with realistic dependencies. \texttt{MCP-Bench} \citep{wangMCPbenchBenchmarkingToolusing2026} represents this shift, connecting 28 live MCP servers exposing 250 tools across finance, travel, scientific computing, and academic search. Its tasks often omit explicit tool names, so the agent must discover relevant tools, pass typed parameters, chain outputs across steps, and ground answers in execution results.

Small language models (1.5B--4B parameters) reduce serving cost, latency, and privacy exposure \citep{belcakSmallLanguageModels2025, erdoganTinyAgentFunctionCalling2024}, but they are brittle in exactly the dimensions that multi-step workflows stress, including malformed JSON, nonexistent tools, missing arguments, broken output chains, and hallucinated answers \citep{patilBerkeleyFunctionCalling2024, gonzalezGorillaLargeLanguage2024, zhangToolBeHonestMultilevelHallucination2024}. 
Figure~\ref{fig:motivation}(a) illustrates the problem. A small planner proposes a workflow that names the right tools but binds incorrect parameters and breaks the dependency chain, producing a plan that looks reasonable yet crashes against live servers. Figure~\ref{fig:motivation}(b) shows how EvoFlux addresses this by executing each candidate, extracting node-level errors, and feeding them back to the same small model as typed edit proposals, maintaining a scored population without any weight updates.

Distillation can help, since SFT teaches output format \citep{chenFireActLanguageAgent2023, qinToolLLMFacilitatingLarge2024} and DPO separates good plans from bad \citep{rafailovDirectPreferenceOptimization2023}, but prior efforts use thousands to tens of thousands of expert trajectories \citep{qinToolLLMFacilitatingLarge2024, zhangXLAMFamilyLarge2025, chenAgentFLANDesigningData2024}, each requiring a teacher call, live execution, and judge evaluation. We operate in a \textbf{scarce-trace} regime of only 177 SFT examples and 624 preference pairs (Appendix~\ref{appendix:dataset}). At this budget, demonstrations teach plan \emph{format} but not the \emph{recovery process}, specifically how to detect wrong tools, fix missing parameters, or repair broken dependencies \citep{luoEmpiricalStudyCatastrophic2025}.

This paper studies the \emph{train-versus-search} tradeoff in the scarce-trace regime. Given the same small planner and the same limited supervision budget, is compute better spent on weight updates (SFT, DPO), or on repairing plans at inference time over the actual task, tool catalog, and execution environment?

\vspace{-1.0em}
\paragraph{Related work.}
Test-time search provides one route around the data bottleneck by spending compute online, through deliberate tree search \citep{yaoTreeThoughtsDeliberate2023}, scaled inference-time compute \citep{snellScalingLLMTesttime2025}, and execution feedback for language agents \citep{shinnReflexionLanguageAgents2023, qiaoMakingLanguageModels2024}. A related evolutionary line treats language models as variation operators that propose, evaluate, and revise structured candidates. FunSearch \citep{romera-paredesMathematicalDiscoveriesProgram2024a} evolves Python functions scored by deterministic evaluators; AlphaEvolve \citep{novikovAlphaEvolveCodingAgent2025} extends this to full programs with frontier-model rewriting; AdaEvolve \citep{cemriAdaEvolveAdaptiveLLM2026} adds a hierarchical controller with bandit-based compute allocation across islands; and FlowEvo \citep{ren2026flowevo} co-evolves workflows and executable skills.

EvoFlux differs from these program-evolution methods in four ways. (1)~It searches over \emph{workflow graphs} whose nodes call live tools through MCP servers, not over source code, so feasibility requires tool resolution, schema validation, and dependency checking against a changing catalog. (2)~It restricts mutation to a finite set of \emph{typed edits} (tool swap, parameter fix, step insertion/removal, reordering, validator insertion) instead of allowing free-form rewriting. (3)~It conditions each mutation on \emph{node-level execution evidence} (e.g., ``node~3 failed: tool not found'') instead of a scalar fitness score alone. (4)~It targets 1.5B--4B planners under a small fixed per-task budget, asking whether scarce demonstrations transfer better through weights or through search-time repair.

We introduce \textbf{EvoFlux}, an inference-time evolutionary search for small tool-using agents. Each plan is a \emph{typed workflow graph}, a DAG whose nodes are tool calls with server, tool name, and parameter bindings, and whose edges carry outputs between steps (Section~\ref{sec:problem}). EvoFlux refines these graphs through typed edits guided by execution errors, with the small model as a proposal operator inside a bounded repair loop. The adaptive intensity and meta-guidance mechanisms adapt the local, within-island controller \citep{cemriAdaEvolveAdaptiveLLM2026} to workflow repair, replacing island populations with action-hash diversity binning over workflow structure and maintaining a fixed per-task budget.

We evaluate five configurations on held-out MCP-Bench tasks, namely zero-shot, SFT, SFT+DPO, ReAct \citep{yaoReActSynergizingReasoning2023}, and EvoFlux. All training baselines use the same \textbf{search-mined} data (Appendix~\ref{appendix:dataset}), testing whether limited demonstrations transfer better through weights or inference-time repair.

The paper makes the following contributions.
\begin{itemize}[leftmargin=*,nosep]
    \item We introduce \textbf{EvoFlux}, an evolutionary procedure that repairs small-model tool plans at inference time using typed workflow edits, execution-grounded scoring, adaptive intensity control, meta-guided redesign, and action-hash diversity pruning.
    \item We frame multi-step tool use by small models as an \emph{executable workflow repair} problem and compare weight updates against inference-time search under the same scarce supervision budget.
    \item Under this budget, inference-time search lifts held-out execution success from ${\sim}3\%$ to 17--24\% across small planners. SFT and SFT+DPO on the same data either match or fall below zero-shot performance. EvoFlux also costs 4--20$\times$ fewer planner tokens per solved task than ReAct.
\end{itemize}

\section{Problem Setup}
\label{sec:problem}
\vspace{-0.5em}
Each task is a triple $q = (x, \mathcal{T}, \mathcal{E})$, consisting of a natural-language request~$x$, a tool catalog~$\mathcal{T}$ listing MCP servers and their tools, and a live execution environment~$\mathcal{E}$. A \emph{workflow planner} emits a \textbf{typed workflow graph}~$g$, a DAG containing (i)~\textbf{tool nodes} binding a server, tool name, parameters, and upstream parents; (ii)~\textbf{dependency edges} carrying outputs between nodes; (iii)~optional \textbf{validator nodes}; and (iv)~a \textbf{terminal output node}. Success requires every tool to exist in~$\mathcal{T}$, every parameter to match the declared schema, every dependency edge to reference a produced output, and the terminal node to answer~$x$ from executed evidence.

Rather than training a planner to emit correct workflows in one pass, we compare five held-out configurations, (1)~zero-shot, (2)~SFT, (3)~SFT+DPO, (4)~ReAct, and (5)~EvoFlux. We partition MCP-Bench into a \emph{search split}~$D_{\mathrm{search}}$ (for development and training-data construction) and a \emph{held-out split}~$D_{\mathrm{eval}}$ (for final comparison only).

\section{Method}
\label{sec:method}
\begin{algorithm}[t]
\caption{Adaptive workflow evolution for one task. Every candidate (initial or mutated) is constructed by \textsc{BuildCandidate} (\S\ref{sec:buildcandidate}). Each line's corresponding paragraph is noted in the margin. Full procedure in Algorithm~\ref{algsearch-detail}.}
\label{algsearch_brief}
\small
\begin{algorithmic}[1]
\Require Task $q$, planner $P$, budget $B$, population cap $K$
\State $c_0 \gets \textsc{InitCandidateWithRetry}(q, P)$ \Comment{Initialization}
\State $\mathcal{P} \gets \{c_0\}$;~ $f^* \gets f(c_0)$;~ $G \gets 0$;~ $\mathrm{cd} \gets 0$
\For{$t = 1$ to $B$}
    \State $I \gets \textsc{Intensity}(G)$ \Comment{Adaptive intensity}
    \If{$G < \tau_m$ and $\mathrm{cd} = 0$}
        \Comment{Meta guidance}
        \State $c_m \gets \textsc{MetaGuidance}(P, q, \mathcal{P})$
        \If{$c_m$ feasible}
            \State $(G, f^*) \gets \textsc{UpdateG}(G, f^*, f(c_m))$
            \State $\mathcal{P} \gets \textsc{Prune}(\mathcal{P} \cup \{c_m\}, K)$
        \EndIf
        \State $\mathrm{cd} \gets \lceil \gamma B \rceil$ \Comment{Cooldown}
    \EndIf
    \State $\mathrm{cd} \gets \max(0, \mathrm{cd} - 1)$
    \State $p \gets \textsc{SelectParent}(\mathcal{P}, I)$ \Comment{Selection}
    \State $e \gets \textsc{CollectEvidence}(p)$ \Comment{Evidence}
    \State $\mathrm{explore} \sim \textsc{Bernoulli}(I)$
    \State $c' \gets \textsc{MutateWithRetry}(P, p, e, \mathrm{explore})$ \Comment{Mutation}
    \If{$\neg c'.\mathrm{feasible}$}
        \State \textbf{continue}
    \EndIf
    \State $(G, f^*) \gets \textsc{UpdateG}(G, f^*, f(c'))$ \Comment{Growth update}
    \State $\mathcal{P} \gets \textsc{Prune}(\mathcal{P} \cup \{c'\}, K)$ \Comment{Pruning}
\EndFor
\State \Return $\mathcal{P}$
\end{algorithmic}
\end{algorithm}

\noindent
\textbf{Overview.}
EvoFlux is an evolutionary loop that repairs tool-use plans at inference time. Given task $q=(x,\mathcal{T},\mathcal{E})$, a planner~$P$ proposes a typed workflow graph (written as JSON). The candidate builder (\textsc{BuildCandidate}, \S\ref{sec:buildcandidate}) compiles this graph by resolving tools against~$\mathcal{T}$, validating parameters against declared schemas, checking dependency structure, and executing when all checks pass. Search then improves the population through typed edits, node-level execution evidence, and planner-guided redesign when mutation stalls. This decomposition suits small models because the planner only needs to propose local improvements to a partially working graph, which demands less global planning capacity than producing a correct workflow in a single pass.

Algorithm~\ref{algsearch_brief} gives the per-task loop; the full procedure appears in Algorithm~\ref{algsearch-detail}. The loop maintains a bounded population~$\mathcal{P}$, the best score~$f^*$, and a growth statistic~$G$ tracking recent improvement. Low~$G$ triggers broader exploration and potential plan redesign; high~$G$ concentrates refinement on the best candidates. Four principles guide the method. First, typed edits ensure every mutation produces a well-formed graph. Second, candidates earn scores through actual execution. Third, each mutation sees node-level error signals. Fourth, search histories yield training data (Appendix~\ref{appendix:dataset}), but held-out execution performance remains the comparison criterion. Implementation details appear in Appendix~\ref{appendix:search-details}.

\begin{figure*}[ht]
    \centering
    \includegraphics[width=0.95\linewidth]{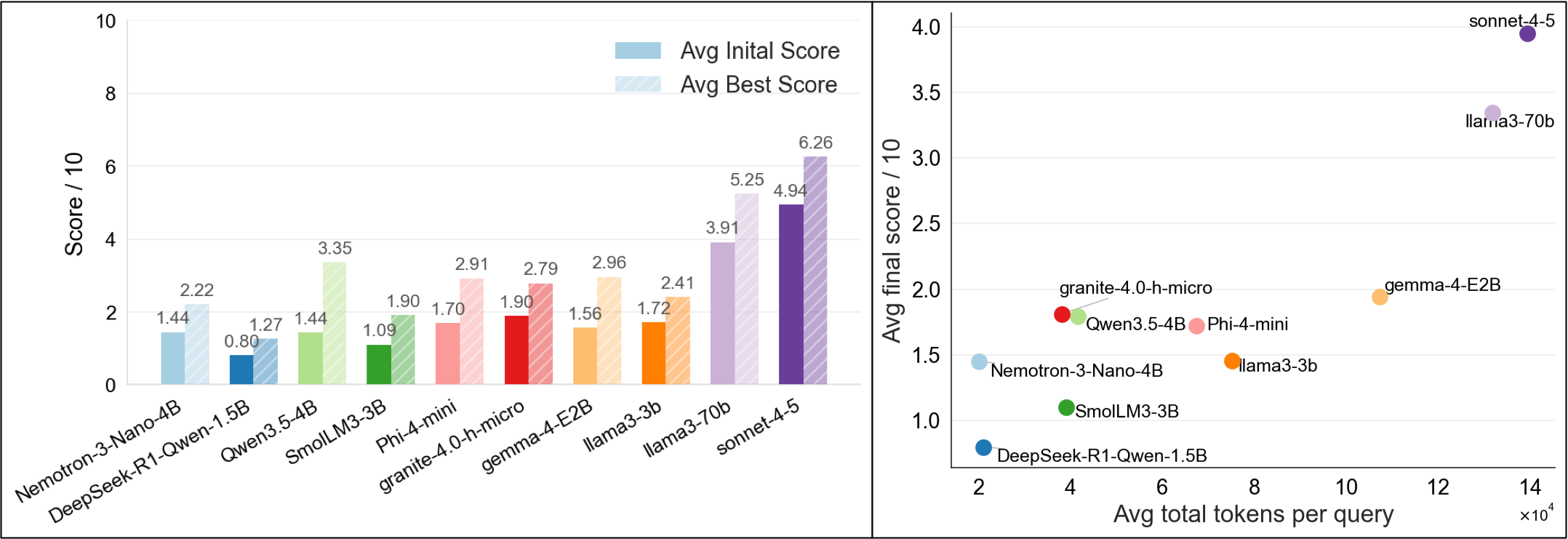}
    \caption{\textbf{Left:} Average initial score vs.\ best score found by EvoFlux per query on the search split (0 = infeasible; 1--10 for feasible candidates; Table~\ref{tab:score_rubric}). \textbf{Right:} Token efficiency frontier, showing average tokens consumed vs.\ delivered score.}
    \label{fig:searchscore_and_seachtoken}
\vspace{-1.5em}
\end{figure*}

\noindent
\textbf{Candidate construction (\textsc{BuildCandidate}).}
\label{sec:buildcandidate}
A candidate~$c$ bundles the typed workflow graph, compiled executable, feasibility metadata, execution logs, and judge scores. \textsc{BuildCandidate} converts a graph into an executable through five stages.
\begin{enumerate}[leftmargin=*,nosep]
    \item \textbf{Resolve} each node's server and tool identifiers against~$\mathcal{T}$ (fail if a tool does not exist).
    \item \textbf{Validate} each node's parameters against the schema declared by the resolved tool (fail if a required field is missing or a type is wrong).
    \item \textbf{Check} graph acyclicity and that every dependency edge refers to a node that executes earlier.
    \item \textbf{Execute} the workflow in~$\mathcal{E}$ when all static checks pass.
\end{enumerate}
Candidates that fail at any stage are marked infeasible and retain the specific error (e.g., ``step~2: tool \texttt{search\_papers} not found on server \texttt{arxiv}''). In the validation task \texttt{context7\_000} (Appendix~\ref{appendix:workflow_structure_analysis}, Figure~\ref{fig_llama_validation_workflows}), the initial graph calls the correct Context7 tools but only weakly connects library resolution to documentation retrieval, failing at stage~4. The repaired graph resolves library identifiers first, feeds them into documentation calls, and routes the retrieved content into a synthesis step, passing all five stages and raising the score from 1.20 to 4.93.

\noindent
\textbf{Initialization.} The planner gets up to $R$~attempts to produce the first feasible workflow, with stage-specific errors returned as context after each failure, equivalent to Reflexion~\citep{shinnReflexionLanguageAgents2023}. Mutation reuses the same retry pattern (Algorithm~\ref{algsearch_brief}, line~17).

\noindent
\textbf{Execution scoring.} The scoring function $f(c)$ uses the MCP-Bench \textsc{LLMJudge} rubric~\citep{wangMCPbenchBenchmarkingToolusing2026}. Infeasible candidates receive $f(c)=0$. For feasible candidates, the judge is invoked five times with shuffled dimension order; each pass rates the six subdimensions in Table~\ref{tab:score_rubric} on a 1--10 scale, and $f(c)$ is their mean. These axes correspond to the failure modes that typed edits target.

\begin{table}[t]
\centering
\footnotesize
\setlength{\tabcolsep}{2pt}
\renewcommand{\arraystretch}{1.1}
\caption{Scoring rubric. Each subdimension is rated 1--10; $f(c)$ is the mean of the six. The rightmost column shows which typed edit addresses failures on that axis.}
\label{tab:score_rubric}
\begin{tabularx}{\columnwidth}{@{}P{0.22\columnwidth}P{0.38\columnwidth}Y@{}}
\toprule
Subdimension & What the judge measures & Repaired by \\
\midrule
Task fulfillment & Fraction of stated task requirements completed & Meta guidance \\
Grounding & Claims supported by tool outputs, not parametric knowledge & Parameter edit, validator insertion \\
Tool appropriateness & Tools well-matched to the task from catalog $\mathcal{T}$ & Tool swap \\
Parameter accuracy & Calls satisfying the tool's declared schema & Parameter edit \\
Dependency awareness & Dependency chains correctly ordered and executed & Step reorder, tool insertion \\
Parallelism \& efficiency & Redundancy and parallel-execution rate & Tool removal \\
\bottomrule
\end{tabularx}
\vspace{-1.75em}
\end{table}

Subdimension scores are logged separately for failure analysis (Tables~\ref{tab:subscore_single}--\ref{tab:subscore_3server}). We use \texttt{openai/gpt-oss-20b} as judge for small models and \texttt{claude/sonnet-4.6} for larger models (Appendix~\ref{appendix:dataset}).

Judge scores drive population management, since parent selection, growth updates, and pruning all rank candidates by $f(c)$. The mutation operator, however, never sees judge outputs. When the planner proposes an edit, its context contains only execution errors from \textsc{CollectEvidence} (e.g., ``tool not found,'' ``schema mismatch,'' ``empty output''), not numeric scores or subdimension ratings. Because the planner sees only concrete execution failures, it cannot optimize for judge preferences; the judge determines which candidates survive, while execution feedback determines how they are repaired. The same judge model, prompt, and five-pass protocol score all five methods on the same held-out tasks.

\noindent
\textbf{Typed edits and execution evidence.} EvoFlux modifies workflows through six typed edit operations.
\begin{enumerate}[leftmargin=*,nosep]
    \item \textbf{Tool swap:} replace a tool binding at a node (fixes ``tool not found'' or poor semantic match).
    \item \textbf{Parameter edit:} revise argument values, add missing fields, or bind an argument to an upstream output (fixes schema errors).
    \item \textbf{Tool insertion:} add a new evidence-gathering or transformation step.
    \item \textbf{Tool removal:} delete a redundant or harmful step and reconnect downstream edges.
    \item \textbf{Step reordering:} change the dependency structure to fix premature calls.
    \item \textbf{Validator insertion:} add a check for malformed or empty intermediate outputs.
\end{enumerate}
Constraining mutation to these operations ensures every mutated graph preserves structural invariants; schema violations are caught at validation and retried. The typed edit space eliminates a large class of malformed outputs that free-form text rewriting would allow.

After each execution, \textsc{CollectEvidence} (Algorithm~\ref{algsearch_brief}, line~15) summarizes what happened at every node. Did the tool resolve? Did parameters match the schema? Did upstream values exist when consumed? Was the output empty or malformed? Did downstream nodes fail because of this node? This produces a per-node error profile that tells the mutation operator \emph{where} and \emph{why} the workflow failed.

\noindent
\textbf{Mutation} (lines~14--17). Each iteration selects a parent~$p$ (line~14, uniformly with probability~$I$ for diversity, otherwise by tournament over high-scoring candidates) and collects its evidence~$e$ (line~15). A Bernoulli draw with probability~$I$ sets the mutation route (line~16).
\begin{itemize}[leftmargin=*,nosep]
    \item \textbf{Exploration} (draw succeeds) applies $n$~stacked random typed edits weighted toward nodes with higher error rates. This path makes \emph{no} planner call; the edits are sampled from the typed edit space rather than proposed by the LLM.
    \item \textbf{Exploitation} (draw fails) calls the planner with a compressed context containing the failing nodes, their neighboring dependencies, the observed errors, the feasible edit operations, and a diff from the current best candidate (capped at 8 nodes). The compressed context omits passing nodes and truncates long tool outputs to keep the prompt within the small model's effective window, so the planner sees the failure site without irrelevant graph regions competing for attention.
\end{itemize}
Both routes produce a typed edit that \textsc{MutateWithRetry} passes to \textsc{BuildCandidate} (line~17). Every candidate clears the same compile $\to$ validate $\to$ execute $\to$ score pipeline.

\noindent
\textbf{Adaptive intensity control} (lines~4--13, 21--22). After each feasible candidate, \textsc{UpdateG} (line~21) computes the nonnegative relative improvement over~$f^*$ and folds its square into an exponential moving average~$G$ (full derivation in Appendix~\ref{appendix:growth_signal}, Equations~1--3). Line~4 converts $G$ into the intensity:
\[
I = I_{\min} + \frac{I_{\max}-I_{\min}}{1+\sqrt{G+\varepsilon}}.
\]
When the population has been improving ($G$ high), $I$ is low, and search concentrates on refining the best candidates. When progress stalls ($G$ low), $I$ rises and search explores more broadly. $I$ drives the two Bernoulli draws described in the mutation paragraph above.

\noindent
\textbf{Meta guidance} (lines~5--11). When $G$ falls below threshold~$\tau_m$, and the cooldown counter is zero, the planner receives the task, the current best candidate, and the three weakest population members, then proposes a \emph{redesigned} workflow. Unlike local mutation, which changes one or two nodes, meta guidance can replace a single-server path with a cross-server workflow, introduce a missing evidence-gathering step, reorganize dependencies, or simplify an overplanned graph. The redesigned candidate passes through the same \textsc{BuildCandidate} pipeline as any other candidate (lines~7--9). A cooldown fraction~$\gamma$ prevents repeated redesign from dominating the loop (line~11), keeping meta guidance as a sparse escape mechanism.

\noindent
\textbf{Selection and pruning} (lines~9, 14, 22). Tournament selection ranks by a Hoeffding lower confidence bound, $\mathrm{LCB} = \bar{f} - \sqrt{\ln(1/\alpha)/(2n)}$, penalizing candidates with few evaluations. The population is capped at $K$~candidates. After each new feasible candidate enters, \textsc{Prune} buckets candidates by a hash of their action structure and selects survivors round-robin across buckets, preserving high-scoring candidates while preventing collapse onto near-duplicate workflows. The final output is the highest-scoring survivor.

\section{Results and Analysis}
\subsection{Search Split}
We first examine the search split to characterize EvoFlux's behavior before testing held-out generalization.

\noindent
\textbf{Initial generation versus search-discovered best.} Figure~\ref{fig:searchscore_and_seachtoken}~\textbf{(left)} compares initial and best-found scores. EvoFlux improves every planner, but the magnitude varies. \texttt{sonnet-4-5} starts at 4.94 and reaches 6.26, gaining 1.32 points (27\%). \texttt{llama3-70b} moves from 3.91 to 5.25, gaining 1.34 points. Among small planners, \texttt{Qwen3.5-4B} shows the largest relative gain at 132\% (1.44$\to$3.35), followed by \texttt{gemma-4-E2B} at 90\% and \texttt{SmolLM3-3B} at 74\%. Search works best when the initial plan contains recoverable structure, amplifying weak but coherent workflows, though it cannot fully rescue planners whose initial graphs provide little useful structure (\texttt{DeepSeek-R1-Distill-Qwen-1.5B} reaches only 1.27). Planners whose initial graphs name at least some relevant tools benefit most, because the evolutionary loop can fix individual nodes without redesigning the entire workflow (see Appendix~\ref{appendix:Search_Subscore} for the per-subskill breakdown).

No sub-4B planner reaches the initial score of \texttt{llama3-70b} or approaches the post-search score of \texttt{sonnet-4-5}. EvoFlux narrows the gap, especially for \texttt{Qwen3.5-4B}, but does not erase the quality advantage of stronger planners.

\begin{figure*}[ht!]
    \centering
    \includegraphics[width=0.98\linewidth]{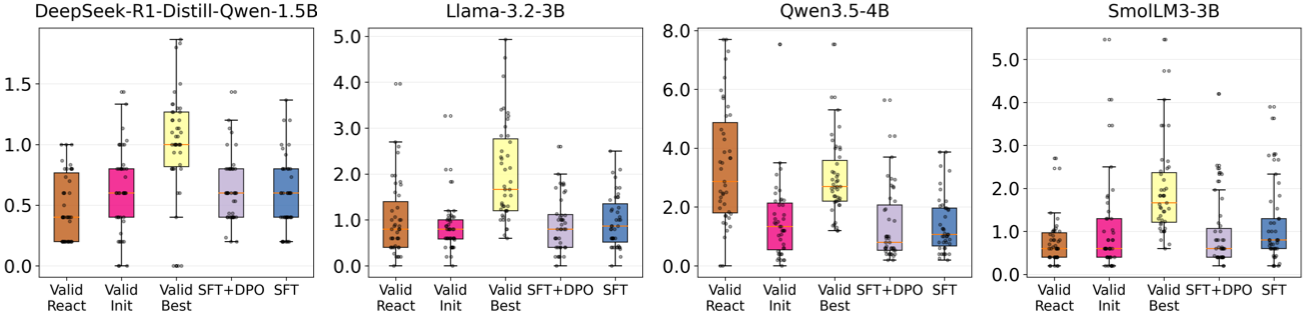}
    \caption{Score distributions (0 = infeasible; 1--10 for feasible candidates; Table~\ref{tab:score_rubric}) across deployment stages for four small planners on the held-out validation split. ``Valid Init'' is the initialization baseline, ``Valid Best'' is the best workflow found by EvoFlux, ``Valid ReAct'' is the ReAct baseline, and ``SFT''/``SFT+DPO'' are the trained checkpoints.}
    \label{fig:validscoredist}
    \vspace{-1.0em}
\end{figure*}

\noindent
\textbf{Token cost versus delivered score.} Figure~\ref{fig:searchscore_and_seachtoken}~(\textbf{right}) compares average token use with the final score. The planners fall into three regimes. \texttt{sonnet-4-5} and \texttt{llama3-70b} occupy the high-cost, high-score region, spending roughly 133--140k tokens per query. An efficient frontier emerges at lower cost: \texttt{Nemotron-3-Nano-4B} uses about 20k tokens for a score near 1.45, while \texttt{granite-4.0-h-micro} and \texttt{Qwen3.5-4B} reach scores around 1.8 at 38--42k tokens.

Several planners sit away from the frontier. \texttt{gemma-4-E2B} spends about 108k tokens for a score near 1.95, less attractive than smaller planners that reach similar scores at much lower cost. This accounting excludes heuristic fallback paths (Appendix~\ref{appendix:heuristic_fallback}). The search split identifies \texttt{Qwen3.5-4B}, \texttt{granite-4.0-h-micro}, and \texttt{Nemotron-3-Nano-4B} as the strongest small-planner tradeoffs; extra token budget alone does not guarantee better workflows.

\begin{figure*}[ht!]
    \centering
    \includegraphics[width=\linewidth]{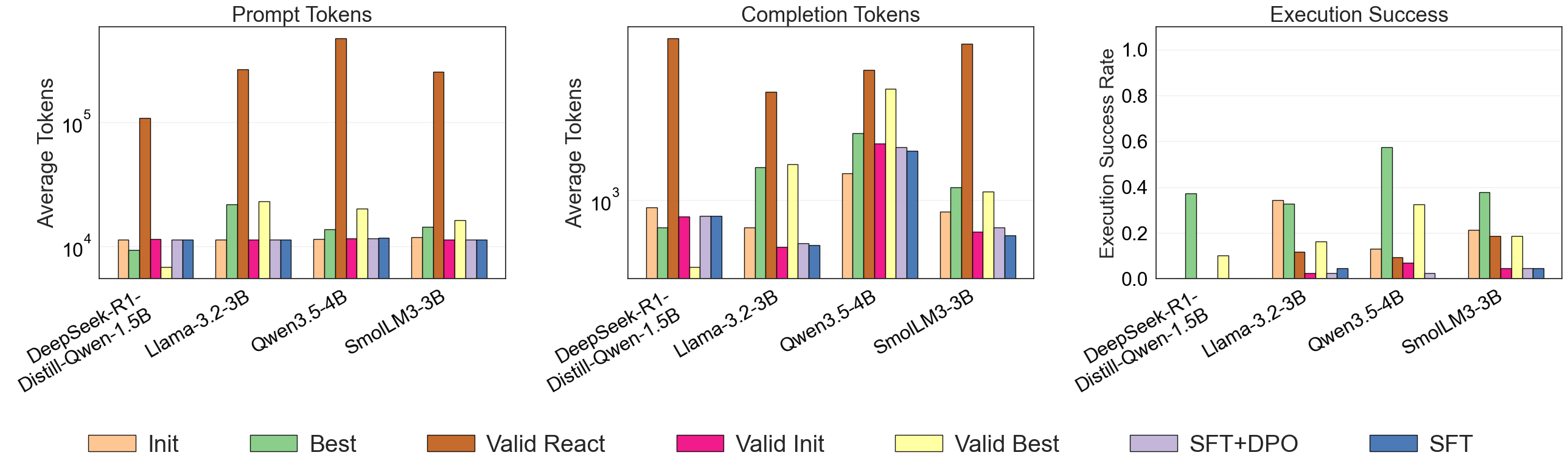}
    \caption{\textbf{Left:} prompt tokens per query, \textbf{Middle:} completion tokens per query, \textbf{Right:} execution success rate (fraction of tasks producing at least one executable workflow). All values shown per deployment stage across the four small planners.}
    \label{fig:validtoken_execution}
    \vspace{-1.0em}
\end{figure*}

\vspace{-0.5em}
\subsection{Validation Split}

The validation split tests whether improvements transfer to held-out tasks across five configurations and four small planners.

\noindent
\textbf{Score distribution across stages.} \label{secvalidexec} Figure~\ref{fig:validscoredist} shows that EvoFlux (Valid Best) consistently shifts scores upward relative to both zero-shot and the trained checkpoints. Execution-guided repair recovers executable workflows that the small models do not learn from 177 training examples. SFT and SFT+DPO remain near zero-shot and sometimes fall below it, indicating that this training budget is insufficient for stable weight-level gains.

ReAct depends more on planner strength. On Qwen3.5-4B (the strongest small planner tested), ReAct achieves the highest scores but with a wider distribution, producing both higher peaks and a longer lower tail. On weaker planners, ReAct provides little improvement over zero-shot because extended tool trajectories amplify drift when the model struggles with multi-step state tracking. EvoFlux yields steadier outcomes with a narrower score spread and a higher floor.

The subskill breakdown (Appendix~\ref{appendix:detailed_subscore}) reveals where EvoFlux's gains concentrate. For Llama-3.2-3B, the single-server overall score rises from 0.912 (Valid Init) to 2.29 (Valid Best), with tool appropriateness climbing from 1.22 to 3.57, parameter accuracy from 1.32 to 3.65, and tool selection from 1.27 to 3.61. SmolLM3-3B follows a similar profile, with three-server parameter accuracy rising from 1.33 to 3.68 and tool selection from 0.988 to 2.86 (full per-model, per-tier breakdowns in Tables~\ref{tab:subscore_single}--\ref{tab:subscore_3server}). Grounding and dependency awareness improve less across all tiers, suggesting that once typed edits fix surface-level tool and parameter errors, the remaining bottleneck shifts to binding correct values and preserving intermediate evidence across steps. 

\noindent
\textbf{Token cost and cost-conditioned comparison.} Figure~\ref{fig:validtoken_execution}~(\textbf{left, middle}) reports prompt and completion \emph{planner} tokens per query, including all intermediate planner calls (mutations, retries, trajectory steps). Judge tokens are excluded because the judge protocol is identical across all five methods. SFT and SFT+DPO are cheapest (a single forward pass), but this economy does not translate into score gains. EvoFlux raises token use through repeated mutation calls, though its cost stays bounded because each mutation conditions on a compact summary of the current candidate and its errors. ReAct dominates token consumption because its trajectory format re-encodes the full history on every step, so prompt tokens grow rapidly with trajectory length.

Table~\ref{tab:cost_outcome} separates costs by outcome. On \emph{solved tasks alone}, ReAct's median cost exceeds EvoFlux's by roughly 20$\times$ on Qwen3.5-4B (370k vs.\ 19k tokens), 4.7$\times$ on Llama-3.2-3B, and 3.2$\times$ on SmolLM3-3B. ReAct's failed runs cost 1.5--10.8$\times$ its successful runs, whereas EvoFlux's failed runs cost 0.7--1.7$\times$ its successes. This asymmetry arises because ReAct can only repair by appending another step to a growing transcript; an early wrong commitment triggers an error spiral up to the step ceiling. EvoFlux instead treats the plan as a revisable object, so when progress stalls, meta guidance abandons local repair and redesigns the graph from compressed failure evidence (Algorithm~\ref{algsearch-detail}, lines~16--23).

Under a fixed deployment budget, the choice between the two methods is a risk-control decision. ReAct can reach higher peaks with a strong planner, but at higher cost and variance. EvoFlux provides more predictable gains at 4--20$\times$ lower planner-token cost per solved task.

\begin{table}[t]
    \centering
    \footnotesize
    \setlength{\tabcolsep}{1.8pt}
    \renewcommand{\arraystretch}{1.08}
    \caption{Token cost conditioned on outcome on the held-out split (43~tasks). Median planner tokens for solved ($f(c) > 0$) and unsolved tasks. Heuristic-fallback runs are excluded from both columns and count as unsolved (score~0) in all other tables.}
    \label{tab:cost_outcome}
    \begin{tabular}{@{}llcrrr@{}}
        \toprule
        Model & Method & \#Sol. & \makecell{Med. tok.\\solved} & \makecell{Med. tok.\\unsolved} & \makecell{U/S\\ratio} \\
        \midrule
        \multirow{2}{*}{Llama-3.2-3B} & ReAct & 5 & 135k & 254k & 1.9$\times$ \\
        & Ours & 7 & 29k & 20k & \textbf{0.7$\times$} \\
        \midrule
        \multirow{2}{*}{Qwen3.5-4B} & ReAct & 4 & 370k & 554k & 1.5$\times$ \\
        & Ours & 14 & 19k & 21k & \textbf{1.1$\times$} \\
        \midrule
        \multirow{2}{*}{\makecell[l]{DeepSeek-R1\\1.5B}} & ReAct & 0 & --- & 111k & --- \\
        & Ours & 3 & 13k & 12k & \textbf{1.0$\times$} \\
        \midrule
        \multirow{2}{*}{SmolLM3-3B} & ReAct & 8 & 34k & 367k & 10.8$\times$ \\
        & Ours & 8 & 11k & 18k & \textbf{1.7$\times$} \\
        \bottomrule
    \end{tabular}
\vspace{-1.0em}
\end{table}

\noindent
\textbf{Execution success across stages.} Figure~\ref{fig:validtoken_execution}~(\textbf{right}) shows a sharp drop in execution success from the search split to held-out tasks, but EvoFlux recovers much of it. For \texttt{Llama-3.2-3B}, zero-shot success falls from ${\sim}35\%$ (search split) to ${\sim}3\%$ (validation), but EvoFlux raises it back to ${\sim}17\%$. For \texttt{Qwen3.5-4B}, zero-shot success drops to ${\sim}3\%$, while EvoFlux reaches ${\sim}24\%$. The trained checkpoints sit at the bottom; SFT and SFT+DPO for \texttt{Qwen3.5-4B} produce essentially no feasible executions (the framework fell back to a heuristic baseline). Training on search-mined traces therefore \emph{reduces} held-out feasibility rather than improving it. This finding aligns with the SFT dataset analysis (Appendix~\ref{appendix:dataset}), where the 177 training examples contain non-oracle positives with mixed execution quality, and the model learns to reproduce common workflow templates without internalizing the dependency logic that makes them executable on unseen tasks.

\noindent
\textbf{Isolating the evolutionary search contribution.} Would simpler strategies such as error-guided retry or best-of-N sampling achieve the same gains without evolutionary search? Three sets of ablations (Appendices~\ref{appendix:ablation}--\ref{appendix:baselines}) address this.

\begin{table}[t]
\centering
\footnotesize
\renewcommand{\arraystretch}{1.1}
\caption{Component ablation on the held-out split (43~tasks). Init+Retry uses up to $R{=}10$ error-guided retries ($\approx$~Reflexion). $\Delta$ columns show the incremental gain from each subsequent component. ``Best from Evo'' reports the fraction of queries whose best score was first achieved during the evolve or meta phase.}
\label{tab:ablation_main}
\resizebox{\columnwidth}{!}{%
\begin{tabular}{@{}l ccc cc c@{}}
\toprule
Model & \makecell{Init+\\Retry} & +Evolve & Full & $\Delta$Evo & $\Delta$Meta & \makecell{Best from\\Evo} \\
\midrule
DeepSeek-1.5B & 0.62 & 1.40 & 1.40 & +0.79 & +0.00 & 88\% \\
Llama-3.2-3B  & 0.87 & 2.29 & 2.45 & +1.42 & +0.16 & 93\% \\
SmolLM3-3B    & 1.13 & 2.09 & 2.15 & +0.96 & +0.07 & 88\% \\
Qwen3.5-4B   & 1.55 & 3.94 & 3.94 & +2.38 & +0.00 & 98\% \\
\bottomrule
\end{tabular}%
}
\end{table}

First, component attribution (Table~\ref{tab:ablation_main}) shows that evolutionary mutations contribute +0.79 to +2.38 points beyond the error-guided retry baseline ($\approx$~Reflexion with $R{=}10$ retries), accounting for 45--60\% of the final score. Across all four planners, 88--98\% of queries achieve their best score during the evolutionary phase. Convergence analysis (Appendix~\ref{appendix:convergence}) confirms that the running best continues to rise throughout the budget without plateauing.

Second, random selection from EvoFlux's own candidate pool at matched candidate counts (Appendix~\ref{appendix:best_of_n}) falls short of guided selection, indicating that evolutionary ordering matters beyond simply generating a diverse pool.

Third, a compute-matched comparison (Appendix~\ref{appendix:compute_matched}) shows that EvoFlux outperforms ReAct on three of four planners while consuming 3--22$\times$ fewer planner tokens. On Qwen3.5-4B, ReAct scores higher but at $9\times$ the token cost.

\noindent
\textbf{Comparison against a frontier planner.} To test whether the repair mechanism generalizes beyond small models, we evaluate Claude-Sonnet-4-6 under the same validation split and rerun small planners under an expanded budget ($B{=}15$, $K{=}15$, \texttt{fb\_repeats}${=}4$, \texttt{max\_steps}${=}25$, \texttt{plan\_retries}${=}10$; Table~\ref{tab_hp_evolution}). Table~\ref{tab:frontier_comparison} shows the results.

\begin{table}[t]
\centering
\caption{Mean score across evaluation stages; subscripts and superscripts denote Q1 and Q3. ``More Resources'' uses the expanded hyperparameters.}
\label{tab:frontier_comparison}
\small
\setlength{\tabcolsep}{5pt}
\renewcommand{\arraystretch}{1.15}
\begin{tabular}{@{}lccc@{}}
\toprule
Model & Valid Init & Valid Best & \makecell{Valid Best\\More Res.} \\
\midrule
Sonnet-4-6 & 2.09$^{2.80}_{0.90}$ & 3.85$^{4.63}_{2.72}$ & n/a \\
Qwen3.5-4B & 1.51$^{2.13}_{0.55}$ & 2.96$^{3.58}_{2.20}$ & 3.94$^{4.95}_{2.68}$ \\
Llama-3.2-3B & 0.82$^{1.00}_{0.58}$ & 2.03$^{2.77}_{1.20}$ & 2.45$^{3.02}_{1.62}$ \\
SmolLM3-3B & 1.04$^{1.30}_{0.40}$ & 1.83$^{2.83}_{0.80}$ & 2.15$^{2.78}_{1.27}$ \\
DeepSeek-1.5B & 0.62$^{0.80}_{0.40}$ & 1.02$^{1.27}_{0.82}$ & 1.40$^{1.57}_{1.20}$ \\
\bottomrule
\end{tabular}
\vspace{-1.0em}
\end{table}

Sonnet itself benefits from the repair loop, rising from 2.09 to 3.85 (an 84\% gain), confirming that execution-grounded repair helps even when the planner already produces reasonable initial workflows. Under the expanded budget, Qwen3.5-4B (4B parameters) reaches 3.94, matching the frontier model's 3.85 with wider variance (Q1--Q3: 2.68--4.95 vs.\ 2.72--4.63). The smaller model compensates for weaker initial proposals with more search iterations, showing that the repair mechanism transfers across model scales.

Llama-3.2-3B and SmolLM3-3B also improve under the expanded budget (reaching 2.45 and 2.15) but remain below the frontier model, consistent with the finding that search amplifies recoverable structure rather than creating it. DeepSeek-1.5B reaches 1.40, reflecting its limited initial generation quality. We detail workflow structure analysis in Appendices~\ref{appendix:depth_width} and~\ref{appendix:workflow_structure_analysis}.

\section{Conclusion}
We frame multi-step tool use by small language models as an executable workflow repair problem and introduce EvoFlux, an inference-time evolutionary search that fixes tool plans through typed edits guided by execution errors. On held-out MCP-Bench tasks, EvoFlux raises execution success from ${\sim}3\%$ to 17--24\% across four small planners (1.5B--4B), while SFT and SFT+DPO on the same 177 examples and 624 preference pairs match or fall below zero-shot performance. ReAct reaches higher peaks on the strongest planner but at 4--20$\times$ the token cost and with higher variance. Under an expanded budget, a 4B planner matches a frontier model's score on this benchmark.

Under scarce supervision, inference compute spent on execution-grounded repair yields more reliable gains than weight updates. Future work should study larger data budgets, learned search controllers, and whether applying EvoFlux on top of trained checkpoints yields complementary gains.

\section*{Limitations}
Our conclusions are scoped to MCP-Bench and the four small planners we evaluate; other benchmarks or deployment environments may shift the tradeoff. The training comparison uses a deliberately small budget (177 SFT examples, 624 preference pairs) and does not test whether larger datasets or reinforcement learning would close the gap with inference-time search. Evaluation relies on LLM-based judging, which we stabilize through five-pass scoring with shuffled dimension order; the judge-independent execution success metric tracks the same trends. We have not tested whether a different judge model or scoring rubric would change the ranking among methods; search selection still depends on the judge. Token costs report only planner tokens because the judge protocol is identical across all five methods.

\section*{Ethical Considerations}
Workflow agents may call tools that access user data, files, APIs, or business systems. Search increases this risk because the agent tries several candidate workflows before selecting one. Deployments should enforce permission checks, sandboxing, rate limits, audit logs, and safeguards around irreversible actions.

We release code and experiment artifacts under the Apache-2.0 license for research on tool-using language agents, execution-grounded workflow search, and evaluation of compact planners. Derived data from MCP-Bench tasks, MCP server outputs, or model-generated execution traces should be used only in ways compatible with the licenses and access terms of the underlying resources.

Before release, we screen prompts, workflow traces, tool outputs, logs, and preference pairs for personally identifiable information (names, email addresses, phone numbers, account identifiers, API keys, and organization-specific secrets). We also screen for offensive or abusive content. Records containing such material are removed or redacted. Released traces are limited to research-safe derived artifacts needed to reproduce the experiments.
\paragraph{Use of AI assistants.}
The study evaluates language model agents within the experimental system. Apart from these experimental agents, AI assistants were used only to refine grammar, wording, and sentence structure in the manuscript. They were not used to design the research questions, create experimental claims, conduct analysis, alter results, annotate data, or make scientific decisions. All technical content, experiments, interpretations, and final writing decisions were reviewed and approved by the authors.

\bibliography{main}

@misc{belcakSmallLanguageModels2025,
  title = {Small {{Language Models}} Are the {{Future}} of {{Agentic AI}}},
  author = {Belcak, Peter and Heinrich, Greg and Diao, Shizhe and Fu, Yonggan and Dong, Xin and Muralidharan, Saurav and Lin, Yingyan Celine and Molchanov, Pavlo},
  year = 2025,
  month = sep,
  number = {arXiv:2506.02153},
  eprint = {2506.02153},
  primaryclass = {cs.AI},
  publisher = {arXiv},
  doi = {10.48550/arXiv.2506.02153},
  urldate = {2026-05-13},
  archiveprefix = {arXiv}
}

@misc{cemriAdaEvolveAdaptiveLLM2026,
  title = {{{AdaEvolve}}: {{Adaptive LLM Driven Zeroth-Order Optimization}}},
  shorttitle = {{{AdaEvolve}}},
  author = {Cemri, Mert and Agrawal, Shubham and Gupta, Akshat and Liu, Shu and Cheng, Audrey and Mang, Qiuyang and Naren, Ashwin and Erdogan, Lutfi Eren and Sen, Koushik and Zaharia, Matei and Dimakis, Alex and Stoica, Ion},
  year = 2026,
  month = feb,
  number = {arXiv:2602.20133},
  eprint = {2602.20133},
  primaryclass = {cs},
  publisher = {arXiv},
  doi = {10.48550/arXiv.2602.20133},
  urldate = {2026-05-10},
  archiveprefix = {arXiv}
}

@inproceedings{erdoganTinyAgentFunctionCalling2024,
  title = {{{TinyAgent}}: {{Function Calling}} at the {{Edge}}},
  shorttitle = {{{TinyAgent}}},
  booktitle = {Proceedings of the 2024 {{Conference}} on {{Empirical Methods}} in {{Natural Language Processing}}: {{System Demonstrations}}},
  author = {Erdogan, Lutfi Eren and Lee, Nicholas and Jha, Siddharth and Kim, Sehoon and Tabrizi, Ryan and Moon, Suhong and Hooper, Coleman Richard Charles and Anumanchipalli, Gopala and Keutzer, Kurt and Gholami, Amir},
  editor = {Hernandez Farias, Delia Irazu and Hope, Tom and Li, Manling},
  year = 2024,
  month = nov,
  pages = {80--88},
  publisher = {Association for Computational Linguistics},
  address = {Miami, Florida, USA},
  doi = {10.18653/v1/2024.emnlp-demo.9},
  urldate = {2026-05-13}
}

@inproceedings{gonzalezGorillaLargeLanguage2024,
  title = {Gorilla: {{Large Language Model Connected}} with {{Massive APIs}}},
  shorttitle = {Gorilla},
  booktitle = {Advances in {{Neural Information Processing Systems}} 37},
  author = {Gonzalez, Joseph and Patil, Shishir and Wang, Xin and Zhang, Tianjun},
  year = 2024,
  pages = {126544--126565},
  publisher = {Neural Information Processing Systems Foundation, Inc. (NeurIPS)},
  address = {Vancouver, BC, Canada},
  doi = {10.52202/079017-4020},
  urldate = {2026-05-10},
  isbn = {979-8-3313-1438-5}
}

@article{luoEmpiricalStudyCatastrophic2025,
  title = {An {{Empirical Study}} of {{Catastrophic Forgetting}} in {{Large Language Models During Continual Fine-Tuning}}},
  author = {Luo, Yun and Yang, Zhen and Meng, Fandong and Li, Yafu and Zhou, Jie and Zhang, Yue},
  year = 2025,
  journal = {IEEE Transactions on Audio, Speech and Language Processing},
  volume = {33},
  pages = {3776--3786},
  issn = {2998-4173},
  doi = {10.1109/TASLPRO.2025.3606231},
  urldate = {2026-05-13}
}

@misc{novikovAlphaEvolveCodingAgent2025,
  title = {{{AlphaEvolve}}: {{A}} Coding Agent for Scientific and Algorithmic Discovery},
  shorttitle = {{{AlphaEvolve}}},
  author = {Novikov, Alexander and V{\~u}, Ng{\^a}n and Eisenberger, Marvin and Dupont, Emilien and Huang, Po-Sen and Wagner, Adam Zsolt and Shirobokov, Sergey and Kozlovskii, Borislav and Ruiz, Francisco J. R. and Mehrabian, Abbas and Kumar, M. Pawan and See, Abigail and Chaudhuri, Swarat and Holland, George and Davies, Alex and Nowozin, Sebastian and Kohli, Pushmeet and Balog, Matej},
  year = 2025,
  month = jun,
  number = {arXiv:2506.13131},
  eprint = {2506.13131},
  primaryclass = {cs},
  publisher = {arXiv},
  doi = {10.48550/arXiv.2506.13131},
  urldate = {2026-05-10},
  archiveprefix = {arXiv}
}

@inproceedings{patilBerkeleyFunctionCalling2024,
  title = 	 {The Berkeley Function Calling Leaderboard ({BFCL}): From Tool Use to Agentic Evaluation of Large Language Models},
  author =       {Patil, Shishir G and Mao, Huanzhi and Yan, Fanjia and Ji, Charlie Cheng-Jie and Suresh, Vishnu and Stoica, Ion and Gonzalez, Joseph E.},
  booktitle = 	 {Proceedings of the 42nd International Conference on Machine Learning},
  pages = 	 {48371--48392},
  year = 	 {2025},
  editor = 	 {Singh, Aarti and Fazel, Maryam and Hsu, Daniel and Lacoste-Julien, Simon and Berkenkamp, Felix and Maharaj, Tegan and Wagstaff, Kiri and Zhu, Jerry},
  volume = 	 {267},
  series = 	 {Proceedings of Machine Learning Research},
  month = 	 {13--19 Jul},
  publisher =    {PMLR},
  url = 	 {https://proceedings.mlr.press/v267/patil25a.html}
}

@inproceedings{qiaoMakingLanguageModels2024,
  title = {Making {{Language Models Better Tool Learners}} with {{Execution Feedback}}},
  booktitle = {Proceedings of the 2024 {{Conference}} of the {{North American Chapter}} of the {{Association}} for {{Computational Linguistics}}: {{Human Language Technologies}} ({{Volume}} 1: {{Long Papers}})},
  author = {Qiao, Shuofei and Gui, Honghao and Lv, Chengfei and Jia, Qianghuai and Chen, Huajun and Zhang, Ningyu},
  editor = {Duh, Kevin and Gomez, Helena and Bethard, Steven},
  year = 2024,
  month = jun,
  pages = {3550--3568},
  publisher = {Association for Computational Linguistics},
  address = {Mexico City, Mexico},
  doi = {10.18653/v1/2024.naacl-long.195},
  urldate = {2026-05-10}
}

@inproceedings{
qinToolLLMFacilitatingLarge2024,
title={Tool{LLM}: Facilitating Large Language Models to Master 16000+ Real-world {API}s},
author={Yujia Qin and Shihao Liang and Yining Ye and Kunlun Zhu and Lan Yan and Yaxi Lu and Yankai Lin and Xin Cong and Xiangru Tang and Bill Qian and Sihan Zhao and Lauren Hong and Runchu Tian and Ruobing Xie and Jie Zhou and Mark Gerstein and dahai li and Zhiyuan Liu and Maosong Sun},
booktitle={The Twelfth International Conference on Learning Representations},
year={2024},
url={https://openreview.net/forum?id=dHng2O0Jjr}
}

@inproceedings{rafailovDirectPreferenceOptimization2023,
  title = {Direct Preference Optimization: Your Language Model Is Secretly a Reward Model},
  shorttitle = {Direct Preference Optimization},
  booktitle = {Proceedings of the 37th {{International Conference}} on {{Neural Information Processing Systems}}},
  author = {Rafailov, Rafael and Sharma, Archit and Mitchell, Eric and Ermon, Stefano and Manning, Christopher D. and Finn, Chelsea},
  year = 2023,
  month = dec,
  series = {{{NIPS}} '23},
  pages = {53728--53741},
  publisher = {Curran Associates Inc.},
  address = {Red Hook, NY, USA},
  urldate = {2026-05-10},
  url={https://dl.acm.org/doi/10.5555/3666122.3668460}
}

@article{romera-paredesMathematicalDiscoveriesProgram2024a,
  title = {Mathematical Discoveries from Program Search with Large Language Models},
  author = {{Romera-Paredes}, Bernardino and Barekatain, Mohammadamin and Novikov, Alexander and Balog, Matej and Kumar, M. Pawan and Dupont, Emilien and Ruiz, Francisco J. R. and Ellenberg, Jordan S. and Wang, Pengming and Fawzi, Omar and Kohli, Pushmeet and Fawzi, Alhussein},
  year = 2024,
  month = jan,
  journal = {Nature},
  volume = {625},
  number = {7995},
  pages = {468--475},
  publisher = {Nature Publishing Group},
  issn = {1476-4687},
  doi = {10.1038/s41586-023-06924-6},
  urldate = {2026-05-10},
  copyright = {2023 The Author(s)},
  langid = {english}
}

@inproceedings{shinnReflexionLanguageAgents2023,
    title={Reflexion: language agents with verbal reinforcement learning},
    author={Noah Shinn and Federico Cassano and Ashwin Gopinath and Karthik R Narasimhan and Shunyu Yao},
    booktitle={Thirty-seventh Conference on Neural Information Processing Systems},
    year={2023},
    url={https://openreview.net/forum?id=vAElhFcKW6}
}

@inproceedings{snellScalingLLMTesttime2025,
    title={Scaling {LLM} Test-Time Compute Optimally Can be More Effective than Scaling Parameters for Reasoning},
    author={Charlie Victor Snell and Jaehoon Lee and Kelvin Xu and Aviral Kumar},
    booktitle={The Thirteenth International Conference on Learning Representations},
    year={2025},
    url={https://openreview.net/forum?id=4FWAwZtd2n}
}

@inproceedings{wangMCPBenchBenchmarkingToolUsing2026,
    title={{MCP}-Bench: Benchmarking Tool-Using {LLM} Agents with Complex Real-World Tasks via {MCP} Servers},
    author={Zhenting Wang and Qi Chang and Hemani Patel and Shashank Biju and Cheng-En Wu and Quan Liu and Aolin Ding and Alireza Rezazadeh and Ankit Shah and Yujia Bao and Eugene Siow},
    booktitle={The Fourteenth International Conference on Learning Representations},
    year={2026},
    url={https://openreview.net/forum?id=fe8mzHwMxN}
}

@inproceedings{yaoTreeThoughtsDeliberate2023,
    title={Tree of Thoughts: Deliberate Problem Solving with Large Language Models},
    author={Shunyu Yao and Dian Yu and Jeffrey Zhao and Izhak Shafran and Thomas L. Griffiths and Yuan Cao and Karthik R Narasimhan},
    booktitle={Thirty-seventh Conference on Neural Information Processing Systems},
    year={2023},
    url={https://openreview.net/forum?id=5Xc1ecxO1h}
}

@inproceedings{zhangToolBeHonestMultilevelHallucination2024,
  title = {{{ToolBeHonest}}: A Multi-Level Hallucination Diagnostic Benchmark for Tool-Augmented Large Language Models},
  booktitle = {Proceedings of the 2024 Conference on Empirical Methods in Natural Language Processing},
  author = {Zhang, Yuxiang and Chen, Jing and Wang, Junjie and Liu, Yaxin and Yang, Cheng and Shi, Chufan and Zhu, Xinyu and Lin, Zihao and Wan, Hanwen and Yang, Yujiu and Sakai, Tetsuya and Feng, Tian and Yamana, Hayato},
  editor = {{Al-Onaizan}, Yaser and Bansal, Mohit and Chen, Yun-Nung},
  year = 2024,
  month = nov,
  pages = {11388--11422},
  publisher = {Association for Computational Linguistics},
  address = {Miami, Florida, USA},
  doi = {10.18653/v1/2024.emnlp-main.637}
}

@misc{chenFireActLanguageAgent2023,
  title = {{{FireAct}}: {{Toward Language Agent Fine-tuning}}},
  shorttitle = {{{FireAct}}},
  author = {Chen, Baian and Shu, Chang and Shareghi, Ehsan and Collier, Nigel and Narasimhan, Karthik and Yao, Shunyu},
  year = 2023,
  month = oct,
  number = {arXiv:2310.05915},
  eprint = {2310.05915},
  primaryclass = {cs.CL},
  publisher = {arXiv},
  doi = {10.48550/arXiv.2310.05915},
  urldate = {2026-05-13},
  archiveprefix = {arXiv}
}

@inproceedings{zhangXLAMFamilyLarge2025,
  title = {{{xLAM}}: A Family of Large Action Models to Empower {{AI}} Agent Systems},
  booktitle = {Proceedings of the 2025 Conference of the Nations of the Americas Chapter of the Association for Computational Linguistics: {{Human}} Language Technologies (Volume 1: {{Long}} Papers)},
  author = {Zhang, Jianguo and Lan, Tian and Zhu, Ming and Liu, Zuxin and Hoang, Thai and Kokane, Shirley and Yao, Weiran and Tan, Juntao and Liu, Zhiwei and Feng, Yihao and Niebles, Juan Carlos and Heinecke, Shelby and Wang, Huan and Savarese, Silvio and Xiong, Caiming},
  editor = {Chiruzzo, Luis and Ritter, Alan and Wang, Lu},
  year = 2025,
  month = apr,
  pages = {11583--11597},
  publisher = {Association for Computational Linguistics},
  address = {Albuquerque, New Mexico},
  doi = {10.18653/v1/2025.naacl-long.578},
  isbn = {979-8-89176-189-6}
}

@inproceedings{chenAgentFLANDesigningData2024,
  title = {Agent-{{FLAN}}: {{Designing}} Data and Methods of Effective Agent Tuning for Large Language Models},
  booktitle = {Findings of the Association for Computational Linguistics: {{ACL}} 2024},
  author = {Chen, Zehui and Liu, Kuikun and Wang, Qiuchen and Zhang, Wenwei and Liu, Jiangning and Lin, Dahua and Chen, Kai and Zhao, Feng},
  editor = {Ku, Lun-Wei and Martins, Andre and Srikumar, Vivek},
  year = 2024,
  month = aug,
  pages = {9354--9366},
  publisher = {Association for Computational Linguistics},
  address = {Bangkok, Thailand},
  doi = {10.18653/v1/2024.findings-acl.557}
}

@inproceedings{yaoReActSynergizingReasoning2023,
    title={ReAct: Synergizing Reasoning and Acting in Language Models},
    author={Shunyu Yao and Jeffrey Zhao and Dian Yu and Nan Du and Izhak Shafran and Karthik R Narasimhan and Yuan Cao},
    booktitle={The Eleventh International Conference on Learning Representations },
    year={2023},
    url={https://openreview.net/forum?id=WE_vluYUL-X}
}

@misc{yue2026static,
      title={From Static Templates to Dynamic Runtime Graphs: A Survey of Workflow Optimization for LLM Agents}, 
      author={Ling Yue and Kushal Raj Bhandari and Ching-Yun Ko and Dhaval Patel and Shuxin Lin and Nianjun Zhou and Jianxi Gao and Pin-Yu Chen and Shaowu Pan},
      year={2026},
      eprint={2603.22386},
      archivePrefix={arXiv},
      primaryClass={cs.AI},
      url={https://arxiv.org/abs/2603.22386}, 
}

@misc{ren2026flowevo,
  title  = {FlowEvo: Self-Evolving Agents through the Co-Evolution of Workflows and Executable Skills},
  author = {Ren, Zeyu and Yue, Ling and Li, Ran and Wang, Yishu and Xu, Shengxiang and Liu, Hanmo and Pan, Shaowu and Di, Shimin},
  year   = {2026},
  doi    = {10.13140/RG.2.2.13642.32960},
  url    = {https://www.researchgate.net/publication/404123514_FlowEvo_Self-Evolving_Agents_through_the_Co-Evolution_of_Workflows_and_Executable_Skills}
}

\clearpage
\newpage 

\appendix

\section{Search implementation details}
\label{appendix:search-details}

\begin{algorithm*}[ht]
\caption{Adaptive workflow evolution for one task, full procedure.}
\label{algsearch-detail}
\small
\begin{algorithmic}[1]
\Require Task $q$, planner $P$, compiler $\mathcal{C}$, executor $\mathcal{E}$,
         budget $B$, population cap $K$, retry budget $R$,
         meta threshold $\tau_m$,
         meta cooldown fraction $\gamma$, EMA factor $\rho$,
         intensity bounds $(I_{\min}, I_{\max})$
\Statex
\State \textbf{Initialization with planner retry}
\State $\mathrm{ctx} \gets \bot$
\For{$r = 1$ to $R$}
    \State $g \gets P.\textsc{plan}(q, \mathrm{ctx})$
    \State $c_0 \gets \textsc{BuildCandidate}(g)$
    \If{$c_0$ feasible and execution succeeded}
        \State \textbf{break}
    \EndIf
    \State $\mathrm{ctx} \gets \textsc{ExtractError}(c_0)$
\EndFor
\State $\mathcal{P} \gets \{c_0\}$;~ $f^* \gets f(c_0)$;~ $G \gets 0$;~ $\mathrm{cd} \gets 0$
\Statex
\State \textbf{Evolutionary loop}
\For{$t = 1$ to $B$}
    \State $I \gets I_{\min} + (I_{\max} - I_{\min}) / (1 + \sqrt{G + \varepsilon})$
    \Statex
    \If{$G < \tau_m$ \textbf{and} $\mathrm{cd} = 0$}
        \Comment{Meta guidance}
        \State $c_m \gets P.\textsc{metaGuide}(q, \mathrm{best}(\mathcal{P}), \mathrm{worst3}(\mathcal{P}))$
        \If{$c_m$ feasible}
            \State $\delta \gets \max\!\big((f(c_m) - f^*) / (|f^*| + \varepsilon),\ 0\big)$
            \State $G \gets \rho G + (1-\rho)\,\delta^2$
            \State $f^* \gets \max(f^*, f(c_m))$
            \State $\mathcal{P} \gets \textsc{Prune}(\mathcal{P} \cup \{c_m\}, K)$
            \State record $c_m$ with phase $= \mathrm{meta}$
        \EndIf
        \State $\mathrm{cd} \gets \lceil \gamma B \rceil$
    \EndIf
    \State $\mathrm{cd} \gets \max(0, \mathrm{cd} - 1)$
    \Statex
    \State $p \gets \textsc{SelectParent}(\mathcal{P}, I)$
    \Comment{uniform w.p. $I$, top-quartile tournament otherwise}
    \State $e \gets \textsc{CollectEvidence}(p)$
    \Comment{per-node error rates from $p$'s execution logs}
    \State $\mathrm{explore} \sim \textsc{Bernoulli}(I)$
    \Statex
    \State $\mathrm{ctx} \gets \bot$;~ $c' \gets \bot$
    \For{$r = 1$ to $R$}
        \Comment{Mutation path with planner retry}
        \If{$\mathrm{explore}$}
            \State $a' \gets \textsc{CompoundRandomEdits}(p, e, n)$
            \Comment{$n$ stacked random edits, evidence-weighted nodes, no LLM}
        \Else
            \State $\nu \gets \textsc{CompressContext}(p, e)$
            \Comment{failing nodes, neighbours, snapshot diff}
            \State $\Omega \gets \textsc{FeasibleOps}(p)$
            \State $a' \gets P.\textsc{proposeEdit}(q, e, \nu, \Omega, \mathrm{ctx})$
        \EndIf
        \State $c' \gets \textsc{BuildCandidate}(a')$
        \If{$c'$ feasible and execution succeeded}
            \State \textbf{break}
        \EndIf
        \State $\mathrm{ctx} \gets \textsc{ExtractError}(c')$
    \EndFor
    \If{$\neg c'.\mathrm{feasible}$}
        \State \textbf{continue}
    \EndIf
    \Statex
    \State $\delta \gets \max\!\big((f(c') - f^*) / (|f^*| + \varepsilon),\ 0\big)$
    \State $G \gets \rho G + (1-\rho)\,\delta^2$
    \State $f^* \gets \max(f^*, f(c'))$
    \State $\mathcal{P} \gets \textsc{Prune}(\mathcal{P} \cup \{c'\}, K)$
    \Comment{16-bucket action-hash binning, round-robin to $K$}
    \State record $c'$ with phase $= \mathrm{evolve}$
\EndFor
\State \Return $\mathcal{P}$
\end{algorithmic}
\end{algorithm*}

\subsection{Action and candidate records}\label{appendix:action_and_candidate}

The implementation separates a symbolic action from a compiled workflow. An action stores the original query, the base graph template, an ordered tuple of typed edits, token cost, prompt parameters, and optional error context. The compiler applies the edits, deduplicates redundant edits, and produces a compiled workflow that can be checked and executed. This separation allows stable action hashes for caching and diversity tracking.

Each candidate stores the action, compiled workflow, score detail, scalar score, feasibility flag, feasibility reasons, execution result, planning attempts, and deterministic candidate identifier. Candidate identifiers derive from query ID, phase, and iteration index. Phases include initialization, evolution, and meta guidance. This layout supports append-only history files and recovery after interruption.

\subsection{Candidate construction}\label{appendix:candidate_detail}

For every query, the planner first produces a base graph with no edits. The system compiles the graph, checks static feasibility, executes feasible workflows, and records the score. If compilation, checking, or execution fails, the system extracts a concise error context and retries planning up to the configured retry budget. This retry loop separates recoverable formatting errors from deeper planning failures.

\subsection{Growth signal and intensity}\label{appendix:growth_signal}

Let $f_t$ be the score of the current child and $f^*_{t-1}$ be the best local score before observing the child. The normalized positive improvement is

\begin{equation}
\delta_t = \max\left( \frac{f_t - f^*_{t-1}}{|f^*_{t-1}| + \epsilon}, 0 \right).
\end{equation}

The accumulated growth signal is an exponential moving average of squared improvements.

\begin{equation}
G_t = \rho G_{t-1} + (1-\rho)\delta_t^2.
\end{equation}

The exploration intensity is

\begin{equation}
I_t = I_{\min} + \frac{I_{\max} - I_{\min}}{1 + \sqrt{G_t + \epsilon}}.
\end{equation}

Intensity controls parent selection and mutation mode. With probability $I_t$, the system samples a parent uniformly from the population. Otherwise, it runs a tournament among high-scoring candidates. With probability $I_t$, mutation enters exploration mode and uses random typed edits instead of an LLM-proposed edit. Low growth, therefore, increases exploration. Strong growth favors exploitation.

\subsection{Heuristic fallback method}
\label{appendix:heuristic_fallback}

The implementation includes a deterministic heuristic fallback for cases in which the planner cannot call, or cannot successfully use, the language model. This fallback is a low-cost safety path that can still return an executable graph template when model-based planning fails. The method deliberately favors guaranteed plan construction over plan quality. It uses only the query text, optional server hints, and the registered tool catalog.

\paragraph{Fallback triggers.}
The \texttt{plan()} method enters the heuristic path under two conditions. First, the method skips the language model immediately when provider credentials are unavailable. If access to the LLM agent is denied, the planner calls the heuristic plan directly. 

\begin{table}[h]
\centering
\small
\begin{tabular}{ll}
\hline
Provider & Required credential \\
\hline
\texttt{bedrock} & Model identifier \\
\texttt{openrouter} & API key \\
\texttt{azure} & API key and endpoint \\
\texttt{base\_url} & Endpoint URL \\
\hline
\end{tabular}
\caption{Provider credential checks used before invoking the model planner.}
\label{tab_heuristic_credentials}
\end{table}

Second, the method enters the heuristic path after repeated language model call failures. Any exception raised during the model call is caught and retried up to \texttt{max\_retries}. With the default setting of \texttt{max\_retries = 2}, the planner makes three total attempts. After the final failed attempt, the planner prints \texttt{[query\_planner] all LLM attempts failed, using heuristic fallback} and calls \texttt{\_heuristic\_plan(query)}.

\paragraph{Fallback algorithm.}
The \texttt{\_heuristic\_plan} uses token overlap between the query and the registered tools. It first lowercases the query text and reads any \texttt{\_servers} hint list from the query metadata. It then tokenizes the query by splitting on non word characters and stores the result as a set of unique query tokens.

For each registered \texttt{(server, tool)} pair, the method concatenates the server name, tool name, and tool description. It tokenizes this combined string with the same non word split and computes the score as the number of overlapping tokens between the query and the tool description string. If the server appears in the \texttt{\_servers} hint list, the method adds a fixed bonus of 10 points. This bonus lets explicit server hints dominate weak lexical overlap while preserving the same scoring machinery for all tools.

The fallback then sorts tools by score in descending order and selects at most \texttt{min(self.max\_steps, 3)} tools. The cap of three steps prevents a failed model call from expanding into a bloated fallback chain. If no tool receives a positive overlap score, the method can still select the highest ranked available tools with score zero. If the registry contains no tools at all, no keyword-based fallback can construct a meaningful tool chain.

The selected tools are assembled into a strictly sequential graph. The first tool has no dependency. The second tool depends on the first. The third tool depends on the second. This linear structure avoids speculative branching and keeps dependency construction simple under fallback conditions.

The method returns a \texttt{GraphTemplate} marked with \texttt{heuristic=True}. The template description is \texttt{heuristic fallback plan}, and \texttt{token\_cost} is set to \texttt{None}, since the path consumes no language model tokens. The heuristic fallback is therefore a deterministic execution floor. It can keep the system moving when credentials are missing or model calls fail, but it cannot reason over schemas, task semantics, or multi-hop evidence with the same flexibility as the model-based planner.

\section{Dataset Splits and Validation}\label{appendix:dataset}

Both datasets are split into training and validation subsets with a validation fraction of $0.12$ and random seed $42$. For SFT, records are shuffled and split directly. For DPO, splitting is performed by query identifier so that all preference pairs for a query remain in the same partition. This avoids leakage in which the model could see one preference pair for a task during training and another during validation.

Validation mirrors the intended role of each dataset. After SFT, we measure JSON syntax rate, schema validity, node coverage against the teacher workflow fingerprint, a score proxy based on node coverage and reference score, a grounding proxy, and a tool appropriateness proxy. These metrics check whether the smaller model has learned to emit parseable workflows that resemble teacher traces and use available tool servers. After DPO, we measure whether the model assigns a higher probability to chosen workflows than rejected workflows on held-out preference pairs, and we rerun the same structural validation used after SFT. This two-level validation separates format learning from preference learning.

The two datasets play different roles. $D_{\mathrm{SFT}}$ teaches the smaller planner the grammar of executable workflows and the common structure of teacher-selected tool graphs. $D_{\mathrm{DPO}}$ sharpens that planner by contrasting higher and lower scoring workflows under the same task context. The statistics in Figures~\ref{fig_sft_dataset} and~\ref{fig_dpo_dataset} also show why training alone may remain insufficient. The SFT set is small and contains non-Oracle positives, while the DPO set contains strong preference gaps but a skewed negative edit distribution. These properties motivate the paper's central comparison between weight-updated smaller planners and inference-time workflow search.

\subsection{SFT and DPO Dataset Construction}

We construct two training datasets from the search split $D_{\mathrm{search}}$. The first dataset supports supervised finetuning, while the second supports direct preference optimization. Both datasets inherit the same workflow representation used by EvoFlux, where each candidate is a structured tool graph with explicit tool nodes, parameters, dependencies, and a terminal output node. This choice keeps the training target aligned with the executable object evaluated at test time, rather than treating tool use as unstructured text generation.

\subsubsection{Supervised Fine Tuning Dataset}
\begin{figure*}[ht]
    \centering
    \includegraphics[width=\textwidth]{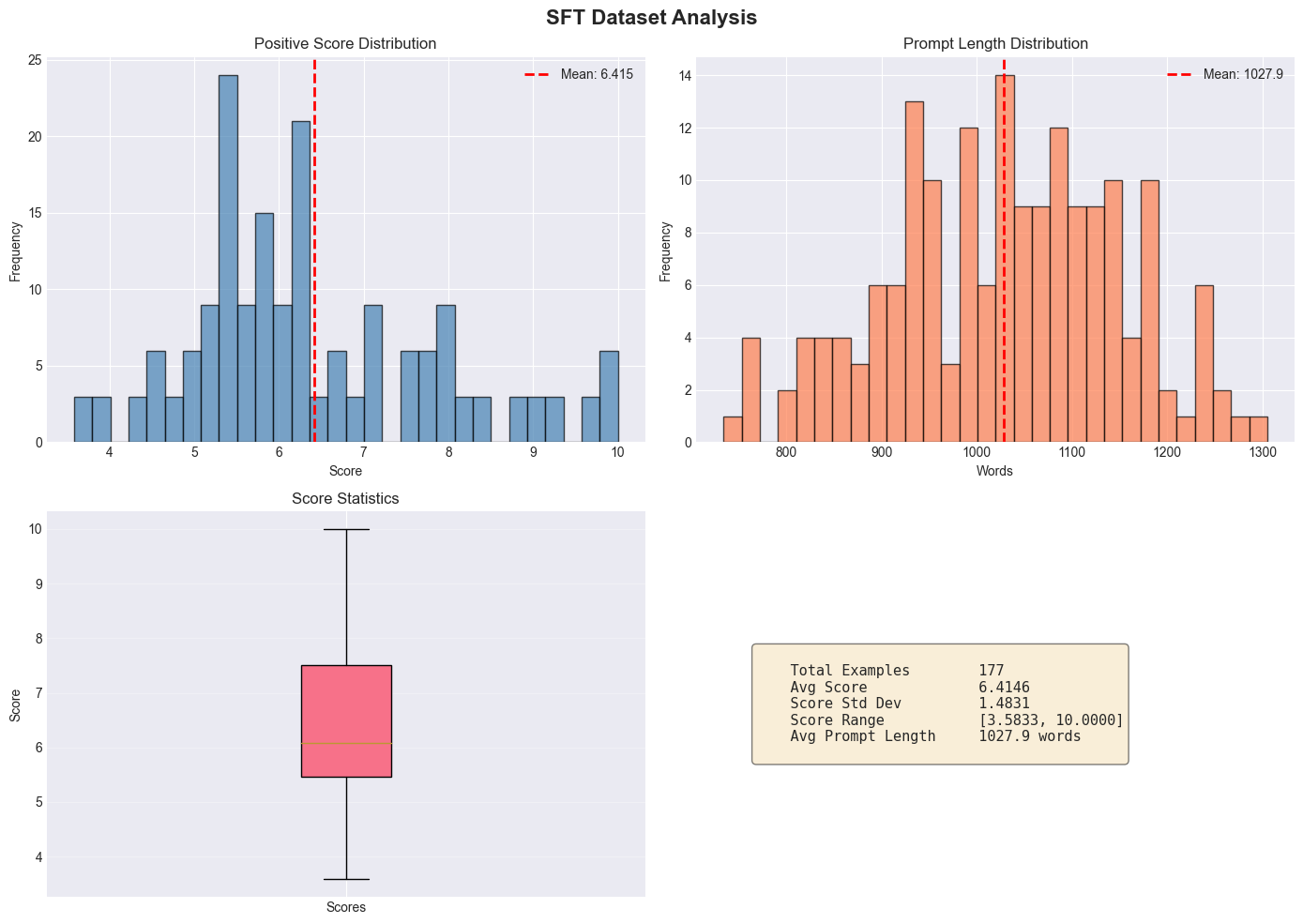}
    \caption{\textbf{SFT dataset diagnostics.} The plots summarize the quality and length profile of the supervised fine tuning data.}
    \label{fig_sft_dataset}
\end{figure*}
The supervised finetuning dataset, denoted $D_{\mathrm{SFT}}$, contains one positive workflow trace for each query. For every task in the search split, we select the highest-scoring workflow generated by the Sonnet teacher and serialize it as the assistant response in a chat-style training example. Each record contains a query identifier, the system and user prompt, the assistant workflow JSON, and the positive execution score assigned to the selected candidate. The training objective, therefore, mimics the teacher's best available executable workflow for each query.

During preprocessing, the system and user messages are formatted as the prompt, while the full conversation, including the assistant workflow, is formatted as the supervised sequence. The completion only collator masks all prompt tokens and computes loss only on the assistant workflow. This design prevents the model from spending capacity memorizing the task and catalog text, and directs the update toward workflow syntax, graph structure, server and tool selection, and node parameter patterns.

The SFT dataset contains $177$ examples. The average positive score is $6.4146$, with a standard deviation of $1.4831$ and a range from $3.5833$ to $10.0000$. The score distribution in Figure~\ref{fig_sft_dataset} concentrates around the mid quality region between roughly $5$ and $7$, while retaining a thinner high quality tail that reaches the maximum score. This distribution shows that the SFT data should be interpreted as behavior cloning from the best-observed teacher candidates, rather than as oracle supervision. Several selected traces remain imperfect under execution-based judging, which matters because a smaller model trained only on this dataset may learn common workflow patterns without fully learning robust repair or search behavior.

The average SFT prompt length is $1027.9$ words. Prompt lengths span roughly $750$ to $1300$ words, with most examples near the mean. This length reflects the information burden of using the MCP-style tool, since each example must expose the task, available servers, tool descriptions, and schema constraints. The box plot of positive scores gives the same picture from a different angle. The median lies near $6$, the interquartile range covers moderate-quality workflows, and the upper whisker extends to perfect executions. SFT therefore provides useful structural supervision, but it has a limited data scale and mixed execution quality.

\subsubsection{DPO Preference Dataset}
\begin{figure*}[t]
    \centering
    \includegraphics[width=\textwidth]{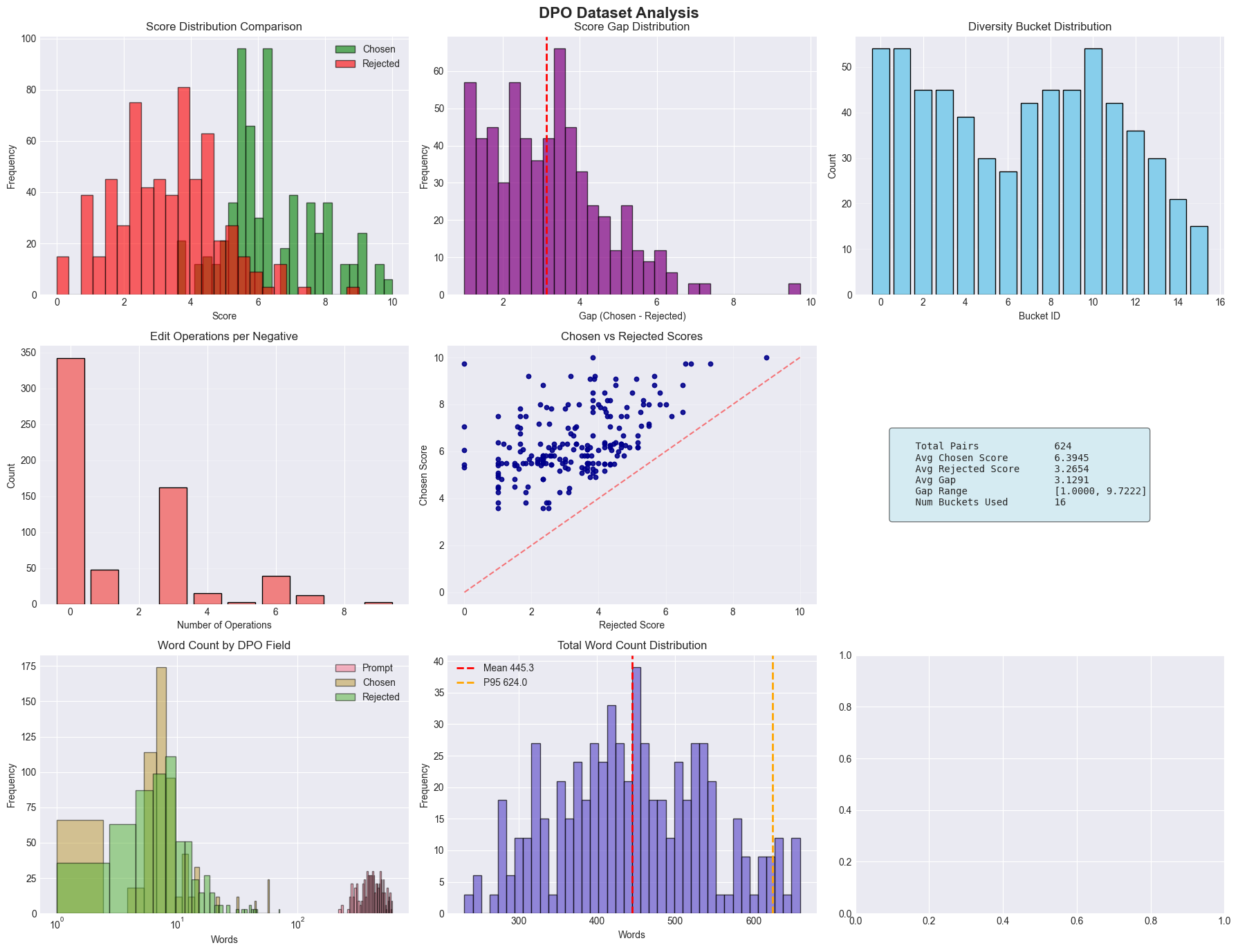}
    \caption{\textbf{DPO dataset analysis.} The plots summarize pairwise preference data constructed from higher and lower scoring workflow candidates. }
    \label{fig_dpo_dataset}
\end{figure*}
The direct preference optimization dataset, denoted $D_{\mathrm{DPO}}$, converts the search trajectories into pairwise preferences. Each DPO record contains a prompt, a chosen workflow, a rejected workflow, a query identifier, the chosen score, the rejected score, and the score gap. The chosen workflow is the higher scoring candidate, while the rejected workflow is a hard negative drawn from lower scoring alternatives for the same task. The DPO stage uses the SFT checkpoint as the reference model, so preference learning adjusts the supervised model toward workflows that score better under execution-based evaluation.

Negative selection uses action identifier bucket diversity, matching the pruning logic used during evolutionary search. Without it, naive preference construction would flood the dataset with near-duplicate failures. Bucketed selection spreads rejected workflows across different action patterns, tool choices, and edit histories. As a result, the preference signal covers a wider set of failure modes, including weak tool selection, missing parameters, poor grounding, broken dependencies, and unnecessary workflow steps.

The DPO dataset contains $624$ preference pairs. The average chosen score is $6.3945$, while the average rejected score is $3.2654$. The mean score gap is $3.1291$, and the gap ranges from $1.0000$ to $9.7222$. Figure~\ref{fig_dpo_dataset} shows that nearly all chosen and rejected pairs fall above the diagonal in the chosen versus rejected scatter plot, confirming that the preference labels encode a consistent ranking signal. The gap histogram also shows a broad spread rather than a narrow margin, which gives DPO both easy separations and harder contrasts.

The score comparison histogram shows substantial separation between chosen and rejected workflows. Chosen candidates cluster mostly between $5$ and $7$, with a tail toward $10$, while rejected candidates concentrate between $1$ and $5$. Some overlap remains in the middle of the score range. That overlap is valuable because it forces the model to distinguish workflows that share superficial structure but differ in execution quality. A preference set with no overlap would mainly teach coarse rejection of broken outputs, while this dataset also teaches finer distinctions among plausible workflows.

The action bucket distribution uses all $16$ buckets. Counts vary across buckets, but no single bucket dominates the dataset, confirming that bucketed sampling spreads the negatives. The rejected edit operation distribution is less balanced. Negatives with zero recorded edit operations dominate the dataset, followed by smaller groups with one, three, six, and a few higher edit counts. This pattern suggests that many rejected candidates come directly from lower-scoring alternatives rather than deep edit chains. The dataset still provides useful contrast, but the skew should be treated as a limitation when interpreting DPO gains.

The DPO field length analysis shows that most training examples remain small enough for stable preference learning. The total word count distribution has a mean of $445.3$ words and a $95$th percentile of $624.0$ words. The prompt field dominates the total length because it carries the task and tool catalog context, while chosen and rejected completions are much shorter serialized workflows. This length profile is well-sized for DPO, where the model compares candidate workflow completions under the same task context.
\newpage
\section{Hyperparameter Configuration}
\label{appendix:hyperparams}

Tables~\ref{tab_hp_evolution}--\ref{tab_hp_lora} report the default hyperparameters used across the five experimental stages. Unless otherwise stated, the same values are used for the search and validation splits.

\begin{table}[t]
\centering
\footnotesize
\setlength{\tabcolsep}{3pt}
\renewcommand{\arraystretch}{1.08}
\caption{EvoFlux hyperparameters used during \texttt{STAGES=search} and \texttt{STAGES=validate}. Perturbations are disabled during validation by setting \texttt{perturbation\_scale=0.0}.}
\label{tab_hp_evolution}
\begin{tabularx}{\columnwidth}{@{}P{0.30\columnwidth}P{0.16\columnwidth}Y@{}}
\toprule
Parameter & Default & Description \\
\midrule
\multicolumn{3}{@{}l}{\textit{Core search budget}} \\
$B$ & 5 & Mutation iterations per query \\
$K$ & 10 & Population cap \\
\hp{fb\_repeats} & 2 & Minibatch repeats \\
\hp{alpha} & 0.05 & LCB confidence level \\
\hp{max\_steps} & 15 & Maximum tool steps per execution \\
\hp{plan\_retries} & 3 & Maximum LLM replanning attempts \\
\hp{seed} & 0 & Random seed \\
\hp{timeout} & 15.0 s & Tool call timeout \\
\midrule
\multicolumn{3}{@{}l}{\textit{Adaptive exploration}} \\
$\rho$ & 0.9 & EMA decay for growth signal $G_t$ \\
$I_{\min}$ & 0.1 & Minimum exploration probability \\
$I_{\max}$ & 0.7 & Maximum exploration probability \\
$\tau_m$ & 0.12 & Meta guidance stagnation threshold \\
\hp{meta\_cooldown\_frac} & 0.2 & Fraction of $B$ between meta guidance calls \\
$\epsilon$ & 1e-8 & Numerical stability constant \\
\midrule
\multicolumn{3}{@{}l}{\textit{Selection and mutation}} \\
Tournament size $k$ & $\min(5,|P|)$ & Contestants per tournament \\
Top quartile size & $|P|/4$ & Exploit pool fraction \\
\hp{max\_context\_nodes} & 8 & Nodes retained in compressed mutation context \\
\hp{robust\_method} & hoeffding & LCB computation method \\
\midrule
\multicolumn{3}{@{}l}{\textit{Search stage perturbations}} \\
\hp{timeout\_inject\_p} & 0.02 & Timeout injection probability \\
\hp{tool\_error\_inject\_p} & 0.02 & Tool error injection probability \\
\hp{latency\_jitter\_ms} & 200 & Maximum latency jitter in milliseconds \\
\bottomrule
\end{tabularx}
\end{table}

\begin{table}[t]
\centering
\footnotesize
\setlength{\tabcolsep}{3pt}
\renewcommand{\arraystretch}{1.08}
\caption{ReAct hyperparameters used during \texttt{STAGES=react}.}
\label{tab_hp_react}
\begin{tabularx}{\columnwidth}{@{}P{0.30\columnwidth}P{0.16\columnwidth}Y@{}}
\toprule
Parameter & Default & Description \\
\midrule
\hp{max\_steps} & 15 & Maximum Thought, Action, Observation rounds \\
\hp{tool\_retries} & 10 & Consecutive tool failures before aborting \\
\hp{parse\_retries} & 3 & Consecutive parse failures before aborting \\
\hp{max\_tokens} & 3096 & Maximum tokens per LLM call \\
\hp{temperature} & 0.0 & Greedy decoding temperature \\
\bottomrule
\end{tabularx}
\end{table}

\begin{table}[t]
\centering
\footnotesize
\setlength{\tabcolsep}{3pt}
\renewcommand{\arraystretch}{1.08}
\caption{SFT hyperparameters used during \texttt{STAGES=sft}.}
\label{tab_hp_sft}
\begin{tabularx}{\columnwidth}{@{}P{0.30\columnwidth}P{0.16\columnwidth}Y@{}}
\toprule
Parameter & Default & Description \\
\midrule
\hp{sft\_epochs} & 2 & Training epochs \\
\hp{sft\_lr} & 1e-5 & Learning rate \\
\hp{sft\_batch} & 1 & Per device batch size \\
\hp{sft\_grad\_accum} & 4 & Gradient accumulation steps \\
\hp{max\_seq\_len} & 2500 & Maximum token sequence length \\
\hp{val\_frac} & 0.12 & Validation holdout fraction \\
\hp{val\_gen} & 20 & Generated examples during validation \\
\hp{gen\_batch} & 2 & Validation generation batch size \\
\hp{lr\_scheduler\_type} & cosine & Learning rate scheduler \\
\hp{eval\_strategy} & epoch & Evaluation frequency \\
\hp{save\_strategy} & epoch & Checkpoint frequency \\
\hp{save\_total\_limit} & 2 & Maximum retained checkpoints \\
\hp{load\_best\_model\_at\_end} & True & Restore best checkpoint after training \\
\hp{metric\_for\_best\_model} & eval loss & Best model selection metric \\
\hp{bf16} & True & BFloat16 precision \\
\hp{tf32} & True & TF32 enabled \\
\hp{gradient\_checkpointing} & True & Memory saving checkpointing \\
\hp{dataloader\_num\_workers} & 0 & Dataloader worker count \\
\hp{seed} & 42 & Train and validation split seed \\
\bottomrule
\end{tabularx}
\end{table}

\begin{table}[t]
\centering
\footnotesize
\setlength{\tabcolsep}{3pt}
\renewcommand{\arraystretch}{1.08}
\caption{DPO hyperparameters used during \texttt{STAGES=sft+dpo}. The reference model is an implicit frozen copy of the policy.}
\label{tab_hp_dpo}
\begin{tabularx}{\columnwidth}{@{}P{0.30\columnwidth}P{0.16\columnwidth}Y@{}}
\toprule
Parameter & Default & Description \\
\midrule
\hp{dpo\_epochs} & 1 & Training epochs \\
\hp{dpo\_lr} & 5e-5 & Learning rate \\
\hp{dpo\_batch} & 1 & Per device batch size \\
\hp{dpo\_grad\_accum} & 4 & Gradient accumulation steps \\
\hp{dpo\_beta} & 0.1 & KL penalty coefficient $\beta$ \\
\hp{dpo\_max\_seq\_len} & 1500 & Maximum token sequence length \\
\hp{dpo\_min\_samples} & 300 & Minimum preference pairs required \\
\hp{val\_dpo\_pairs} & 40 & Maximum pairs used for reward margin evaluation \\
\hp{warmup\_ratio} & 0.1 & Learning rate warmup fraction \\
\hp{lr\_scheduler\_type} & cosine & Learning rate scheduler \\
\hp{bf16} & True & BFloat16 precision \\
\hp{tf32} & True & TF32 enabled \\
\hp{gradient\_checkpointing} & False & Disabled for DPO \\
\hp{save\_total\_limit} & 2 & Maximum retained checkpoints \\
\hp{load\_best\_model\_at\_end} & True & Restore best checkpoint after training \\
\hp{metric\_for\_best\_model} & eval loss & Best model selection metric \\
\hp{truncation\_mode} & keep end & Sequence truncation strategy \\
\hp{precompute\_ref\_log\_probs} & True & Precompute reference log probabilities \\
\hp{precompute\_ref\_batch\_size} & 1 & Reference precompute batch size \\
\bottomrule
\end{tabularx}
\end{table}

\begin{table}[t]
\centering
\footnotesize
\setlength{\tabcolsep}{3pt}
\renewcommand{\arraystretch}{1.08}
\caption{Shared LoRA adapter configuration used for both SFT and DPO. Quantization is disabled by default.}
\label{tab_hp_lora}
\begin{tabularx}{\columnwidth}{@{}P{0.30\columnwidth}P{0.16\columnwidth}Y@{}}
\toprule
Parameter & Default & Description \\
\midrule
\hp{lora\_r} & 16 & LoRA rank \\
\hp{lora\_alpha} & 32 & LoRA scaling parameter \\
\hp{lora\_dropout} & 0.1 & Dropout rate \\
\hp{lora\_bias} & none & Bias adaptation setting \\
\hp{lora\_task\_type} & CAUSAL LM & Causal LM fine tuning objective \\
\hp{target\_modules} & listed & \hp{q\_proj}, \hp{k\_proj}, \hp{v\_proj}, \hp{o\_proj}, \hp{gate\_proj}, \hp{up\_proj}, \hp{down\_proj} \\
\bottomrule
\end{tabularx}
\end{table}

\newpage
\section{Subskill Comparison}
\subsection{Subskill profile before and after search on the Search Split}\label{appendix:Search_Subscore}
\begin{figure*}[t]
    \centering
    \includegraphics[width=0.98\linewidth]{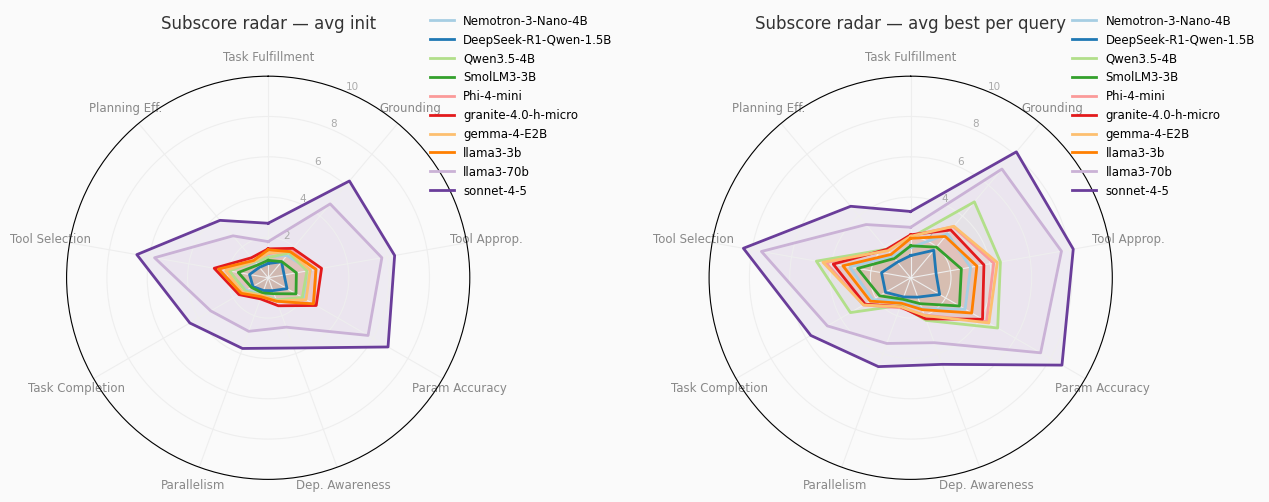}
    \caption{Subskill radar plots before and after search on the search split}
    \label{fig:searchsubscore}
\end{figure*}

Figure~\ref{fig:searchsubscore} shows how evolution search changes the six judge dimensions for different models. Before search, \texttt{sonnet-4-5} and \texttt{llama3-70b} dominate the radar plot across nearly all axes. Smaller planners cluster near the center, with most subskill scores around 1 to 2. After the search, the smaller profiles expand, but the gains concentrate in specific dimensions.

The largest improvements appear in tool selection, tool appropriateness, and task completion. \texttt{Qwen3.5-4B} improves on tool appropriateness enough to approach the \texttt{llama3-70b} contour on that axis. Typed edits can repair semantic tool-matching errors when the planner has already proposed a plausible workflow skeleton. Search also raises task completion for several smaller planners, indicating that execution feedback helps convert partially correct workflows into more useful final answers.

Grounding and parameter accuracy improve less. Even after search, smaller planners stay close to the center on these axes, while \texttt{sonnet-4-5} expands on grounding and tool appropriateness. Dependency awareness and parallelism improve modestly across planners. After search, the bottleneck for smaller agents shifts from finding a plausible tool to binding correct values, preserving intermediate evidence, and feeding downstream calls with the right arguments.

\newpage
\subsection{Subskill profile before and after search on the Validation Split}\label{appendix:detailed_subscore}

\begin{table*}[t]
\centering
\footnotesize
\setlength{\tabcolsep}{2.5pt}
\renewcommand{\arraystretch}{1.15}
\caption{Subscore performance on \textbf{single-server} tasks (held-out split). Scores: 0 = infeasible; 1--10 for feasible candidates.}
\label{tab:subscore_single}
\begin{tabular}{@{}llcccccccccc@{}}
\toprule
Model & Stage & \textbf{Overall} & \makecell{Task\\fulfill.} & Ground. & \makecell{Tool\\appropr.} & \makecell{Param.\\acc.} & \makecell{Dep.\\aware.} & \makecell{Par.\\eff.} & \makecell{Task\\compl.} & \makecell{Tool\\sel.} & \makecell{Plan.\\eff.} \\
\midrule
\multirow{5}{*}{\makecell[l]{DeepSeek-R1-\\Distill-Qwen-1.5B}}
 & Valid ReAct & 0.554 & 0.548 & 0.626 & 0.548 & 0.539 & 0.53 & 0.53 & 0.587 & 0.543 & 0.53 \\
 & Valid Init & 0.601 & 0.6 & 0.733 & 0.771 & 0.676 & 0.59 & 0.581 & 0.667 & 0.724 & 0.586 \\
 & SFT & 0.496 & 0.496 & 0.496 & 0.496 & 0.496 & 0.496 & 0.496 & 0.496 & 0.496 & 0.496 \\
 & SFT+DPO & 0.68 & 0.643 & 0.852 & 0.626 & 0.704 & 0.626 & 0.626 & 0.748 & 0.665 & 0.626 \\
 & Valid Best & 0.981 & 0.895 & 1.37 & 1.05 & 1.42 & 0.867 & 0.848 & 1.13 & 1.23 & 0.857 \\
\midrule
\multirow{5}{*}{Llama-3.2-3B}
 & Valid ReAct & 0.906 & 0.748 & 1.24 & 1.1 & 0.913 & 0.765 & 0.661 & 0.996 & 1.01 & 0.713 \\
 & Valid Init & 0.912 & 0.739 & 0.835 & 1.22 & 1.32 & 0.678 & 0.678 & 0.787 & 1.27 & 0.678 \\
 & SFT & 0.951 & 0.704 & 0.93 & 1.2 & 1.5 & 0.704 & 0.67 & 0.817 & 1.35 & 0.687 \\
 & SFT+DPO & 0.855 & 0.652 & 1.1 & 0.939 & 1.17 & 0.643 & 0.626 & 0.874 & 1.06 & 0.635 \\
 & Valid Best & 2.29 & 1.57 & 2.37 & 3.57 & 3.65 & 1.49 & 1.06 & 1.97 & 3.61 & 1.27 \\
\midrule
\multirow{5}{*}{Qwen3.5-4B}
 & Valid ReAct & 3.54 & 2.5 & 3.9 & 5.05 & 5.24 & 2.9 & 1.65 & 3.2 & 5.14 & 2.28 \\
 & Valid Init & 1.82 & 1.23 & 2.04 & 2.63 & 2.7 & 1.44 & 0.896 & 1.63 & 2.67 & 1.17 \\
 & SFT & 1.28 & 0.828 & 1.47 & 1.68 & 1.96 & 0.898 & 0.872 & 1.15 & 1.82 & 0.885 \\
 & SFT+DPO & 1.57 & 0.952 & 1.69 & 2.49 & 2.52 & 1.03 & 0.748 & 1.32 & 2.51 & 0.888 \\
 & Valid Best & 3.2 & 2.07 & 3.84 & 4.73 & 4.95 & 2.31 & 1.29 & 2.96 & 4.84 & 1.8 \\
\midrule
\multirow{5}{*}{SmolLM3-3B}
 & Valid ReAct & 0.751 & 0.652 & 0.983 & 0.861 & 0.791 & 0.609 & 0.609 & 0.817 & 0.826 & 0.609 \\
 & Valid Init & 1.37 & 0.887 & 1.41 & 1.89 & 2.36 & 0.948 & 0.722 & 1.15 & 2.12 & 0.835 \\
 & SFT & 1.28 & 0.861 & 1.43 & 1.88 & 1.89 & 0.887 & 0.713 & 1.15 & 1.88 & 0.8 \\
 & SFT+DPO & 0.997 & 0.809 & 1.1 & 1.34 & 1.28 & 0.73 & 0.722 & 0.957 & 1.31 & 0.726 \\
 & Valid Best & 2.03 & 1.36 & 2.48 & 2.59 & 3.43 & 1.3 & 1.02 & 1.92 & 3.01 & 1.16 \\
\bottomrule
\end{tabular}
\end{table*}

\begin{table*}[t]
\centering
\footnotesize
\setlength{\tabcolsep}{2.5pt}
\renewcommand{\arraystretch}{1.15}
\caption{Subscore performance on \textbf{2-server} tasks (held-out split).}
\label{tab:subscore_2server}
\begin{tabular}{@{}llcccccccccc@{}}
\toprule
Model & Stage & \textbf{Overall} & \makecell{Task\\fulfill.} & Ground. & \makecell{Tool\\appropr.} & \makecell{Param.\\acc.} & \makecell{Dep.\\aware.} & \makecell{Par.\\eff.} & \makecell{Task\\compl.} & \makecell{Tool\\sel.} & \makecell{Plan.\\eff.} \\
\midrule
\multirow{5}{*}{\makecell[l]{DeepSeek-R1-\\Distill-Qwen-1.5B}}
 & Valid ReAct & 0.464 & 0.433 & 0.6 & 0.45 & 0.433 & 0.433 & 0.433 & 0.517 & 0.442 & 0.433 \\
 & Valid Init & 0.692 & 0.727 & 0.855 & 0.709 & 0.891 & 0.673 & 0.673 & 0.791 & 0.8 & 0.673 \\
 & SFT & 0.649 & 0.458 & 1.28 & 0.625 & 0.608 & 0.458 & 0.458 & 0.871 & 0.617 & 0.458 \\
 & SFT+DPO & 0.6 & 0.6 & 0.733 & 0.567 & 0.567 & 0.567 & 0.567 & 0.667 & 0.567 & 0.567 \\
 & Valid Best & 1.0 & 1.05 & 1.33 & 1.02 & 1.16 & 1.0 & 1.0 & 1.19 & 1.09 & 1.0 \\
\midrule
\multirow{5}{*}{Llama-3.2-3B}
 & Valid ReAct & 1.49 & 1.25 & 1.65 & 1.82 & 1.82 & 1.45 & 0.95 & 1.45 & 1.82 & 1.2 \\
 & Valid Init & 0.725 & 0.55 & 0.85 & 0.917 & 0.867 & 0.583 & 0.583 & 0.7 & 0.892 & 0.583 \\
 & SFT & 0.956 & 0.767 & 1.45 & 1.2 & 0.8 & 0.8 & 0.717 & 1.11 & 1.0 & 0.758 \\
 & SFT+DPO & 0.956 & 0.717 & 1.02 & 1.28 & 1.23 & 0.733 & 0.75 & 0.867 & 1.26 & 0.742 \\
 & Valid Best & 1.91 & 1.32 & 2.43 & 2.62 & 2.85 & 1.17 & 1.1 & 1.88 & 2.73 & 1.13 \\
\midrule
\multirow{5}{*}{Qwen3.5-4B}
 & Valid ReAct & 2.85 & 2.4 & 3.58 & 3.42 & 3.78 & 2.48 & 1.45 & 2.99 & 3.6 & 1.97 \\
 & Valid Init & 1.0 & 0.737 & 1.35 & 1.09 & 1.6 & 0.671 & 0.554 & 1.05 & 1.35 & 0.612 \\
 & SFT & 1.57 & 0.946 & 1.53 & 2.01 & 2.95 & 1.05 & 0.929 & 1.24 & 2.48 & 0.987 \\
 & SFT+DPO & 1.47 & 0.933 & 1.37 & 2.33 & 2.48 & 0.9 & 0.817 & 1.15 & 2.41 & 0.858 \\
 & Valid Best & 2.34 & 1.97 & 2.32 & 3.17 & 3.75 & 1.43 & 1.38 & 2.14 & 3.46 & 1.41 \\
\midrule
\multirow{5}{*}{SmolLM3-3B}
 & Valid ReAct & 0.553 & 0.567 & 0.55 & 0.55 & 0.55 & 0.55 & 0.55 & 0.558 & 0.55 & 0.55 \\
 & Valid Init & 0.653 & 0.65 & 0.867 & 0.617 & 0.633 & 0.583 & 0.567 & 0.758 & 0.625 & 0.575 \\
 & SFT & 0.892 & 0.675 & 0.771 & 0.938 & 1.7 & 0.638 & 0.638 & 0.723 & 1.32 & 0.638 \\
 & SFT+DPO & 0.917 & 0.817 & 1.05 & 0.983 & 1.13 & 0.767 & 0.75 & 0.933 & 1.06 & 0.758 \\
 & Valid Best & 1.85 & 1.48 & 2.18 & 2.44 & 2.51 & 1.41 & 1.07 & 1.83 & 2.48 & 1.24 \\
\bottomrule
\end{tabular}
\end{table*}

\begin{table*}[t]
\centering
\footnotesize
\setlength{\tabcolsep}{2.5pt}
\renewcommand{\arraystretch}{1.15}
\caption{Subscore performance on \textbf{3-server} tasks (held-out split).}
\label{tab:subscore_3server}
\begin{tabular}{@{}llcccccccccc@{}}
\toprule
Model & Stage & \textbf{Overall} & \makecell{Task\\fulfill.} & Ground. & \makecell{Tool\\appropr.} & \makecell{Param.\\acc.} & \makecell{Dep.\\aware.} & \makecell{Par.\\eff.} & \makecell{Task\\compl.} & \makecell{Tool\\sel.} & \makecell{Plan.\\eff.} \\
\midrule
\multirow{5}{*}{\makecell[l]{DeepSeek-R1-\\Distill-Qwen-1.5B}}
 & Valid ReAct & 0.337 & 0.35 & 0.35 & 0.325 & 0.325 & 0.35 & 0.325 & 0.35 & 0.325 & 0.338 \\
 & Valid Init & 0.575 & 0.575 & 0.575 & 0.575 & 0.575 & 0.575 & 0.575 & 0.575 & 0.575 & 0.575 \\
 & SFT & 0.725 & 0.725 & 0.725 & 0.725 & 0.725 & 0.725 & 0.725 & 0.725 & 0.725 & 0.725 \\
 & SFT+DPO & 0.554 & 0.575 & 0.55 & 0.55 & 0.55 & 0.55 & 0.55 & 0.562 & 0.55 & 0.55 \\
 & Valid Best & 1.18 & 1.15 & 1.7 & 1.05 & 1.05 & 1.05 & 1.05 & 1.43 & 1.05 & 1.05 \\
\midrule
\multirow{5}{*}{Llama-3.2-3B}
 & Valid ReAct & 0.662 & 0.625 & 0.75 & 0.8 & 0.7 & 0.55 & 0.55 & 0.688 & 0.75 & 0.55 \\
 & Valid Init & 0.683 & 0.6 & 0.6 & 1.07 & 0.6 & 0.625 & 0.6 & 0.6 & 0.838 & 0.613 \\
 & SFT & 1.01 & 0.825 & 1.0 & 1.52 & 1.18 & 0.8 & 0.725 & 0.912 & 1.35 & 0.762 \\
 & SFT+DPO & 1.06 & 0.7 & 0.85 & 2.15 & 1.3 & 0.7 & 0.675 & 0.775 & 1.73 & 0.688 \\
 & Valid Best & 1.45 & 1.32 & 1.57 & 1.68 & 1.9 & 1.12 & 1.1 & 1.45 & 1.79 & 1.11 \\
\midrule
\multirow{5}{*}{Qwen3.5-4B}
 & Valid ReAct & 3.95 & 3.48 & 4.5 & 5.1 & 5.15 & 3.8 & 1.7 & 3.99 & 5.12 & 2.75 \\
 & Valid Init & 1.36 & 1.15 & 0.975 & 2.07 & 2.09 & 0.969 & 0.919 & 1.06 & 2.08 & 0.944 \\
 & SFT & 1.42 & 0.9 & 1.55 & 2.17 & 2.38 & 0.775 & 0.725 & 1.23 & 2.27 & 0.75 \\
 & SFT+DPO & 0.683 & 0.55 & 0.425 & 1.27 & 0.9 & 0.475 & 0.475 & 0.488 & 1.09 & 0.475 \\
 & Valid Best & 3.2 & 2.2 & 4.4 & 3.88 & 5.3 & 2.02 & 1.4 & 3.3 & 4.59 & 1.71 \\
\midrule
\multirow{5}{*}{SmolLM3-3B}
 & Valid ReAct & 1.03 & 0.875 & 1.45 & 0.825 & 1.4 & 0.825 & 0.8 & 1.16 & 1.11 & 0.812 \\
 & Valid Init & 0.654 & 0.488 & 0.488 & 0.644 & 1.33 & 0.488 & 0.488 & 0.488 & 0.988 & 0.488 \\
 & SFT & 0.958 & 0.85 & 0.75 & 1.12 & 1.52 & 0.75 & 0.75 & 0.8 & 1.32 & 0.75 \\
 & SFT+DPO & 0.883 & 0.75 & 0.675 & 0.775 & 1.72 & 0.7 & 0.675 & 0.713 & 1.25 & 0.688 \\
 & Valid Best & 1.85 & 1.34 & 1.66 & 2.04 & 3.68 & 1.14 & 1.24 & 1.5 & 2.86 & 1.19 \\
\bottomrule
\end{tabular}
\end{table*}

Tables~\ref{tab:subscore_single}--\ref{tab:subscore_3server} shows that EvoFlux improves small agents by repairing specific mechanics of executable tool use. The largest gains for Llama-3.2-3B and SmolLM3-3B cluster around tool selection, tool appropriateness, parameter accuracy, and grounding, the failure modes targeted by typed workflow edits and execution feedback. For Llama-3.2-3B, Valid Best raises the single-server overall score from 0.912 to 2.29: tool appropriateness climbs from 1.22 to 3.57, parameter accuracy from 1.32 to 3.65, grounding from 0.835 to 2.37, and tool selection from 1.27 to 3.61. The same pattern appears at two servers, where the overall score rises from 0.725 to 1.91. SmolLM3-3B follows the same profile, with overall scores rising from 1.37 to 2.03 at one server, 0.653 to 1.85 at two servers, and 0.654 to 1.85 at three servers. At three servers, parameter accuracy rises from 1.33 to 3.68 and tool selection from 0.988 to 2.86.

Qwen3.5-4B shows a different pattern. ReAct posts the highest overall scores for Qwen at every complexity tier, beating Valid Best by 0.34 points on single-server tasks, 0.51 on two-server tasks, and 0.75 on three-server tasks. Across 3 servers, ReAct scores 3.95 overall vs.\ 3.20 for Valid Best, 5.10 for tool appropriateness vs.\ 3.88, and 5.12 for tool selection vs.\ 4.59. Parameter accuracy is the one exception: ReAct wins at one server (5.24 vs.\ 4.95), nearly ties at two servers (3.78 vs.\ 3.75), and trails Valid Best at three servers (5.15 vs.\ 5.30). ReAct rewards Qwen when long tool trajectories remain coherent. EvoFlux supplies controlled repair when trajectory coherence breaks down, delivering more predictable gains under a lower token budget.

Training results are mixed. Llama-3.2-3B's trained checkpoints lift overall scores above Valid Init at two and three servers (SFT and SFT+DPO both reaching 0.956 at two servers vs.\ 0.725 for Valid Init, and 1.01 and 1.06 at three servers vs.\ 0.683), but both remain below Valid Best at every tier (2.29, 1.91, and 1.45). Qwen3.5-4B shows a sharper failure: SFT+DPO falls to 0.683 overall at three servers, well below Valid Init at 1.36 and Valid Best at 3.20. Grounding drops from 0.975 at Valid Init to 0.425 after SFT+DPO; tool selection drops from 2.08 to 1.09. Finetuning under a small trace budget behaves like shallow imitation of workflow form, learning some tool-call shapes while failing to internalize the repair process that produced successful workflows.

\paragraph{Failure mode distribution across complexity tiers.} The subscore tables show how EvoFlux's gains distribute across dimensions as task complexity increases. At the single-server level (Table~\ref{tab:subscore_single}), the largest absolute improvements for Llama-3.2-3B concentrate in tool appropriateness ($+2.35$, from 1.22 to 3.57), tool selection ($+2.34$, from 1.27 to 3.61), and parameter accuracy ($+2.33$, from 1.32 to 3.65). SmolLM3-3B follows a similar profile, with parameter accuracy gaining $+1.07$ (2.36 to 3.43) and tool appropriateness $+0.70$ (1.89 to 2.59). At the two-server level (Table~\ref{tab:subscore_2server}), parameter accuracy becomes the largest single gain for both Llama-3.2-3B ($+1.98$, from 0.867 to 2.85) and SmolLM3-3B ($+1.88$, from 0.633 to 2.51), overtaking tool appropriateness ($+1.70$ and $+1.82$ respectively). Cross-server workflows require binding outputs from one server's tools as inputs to another's, and small models frequently pass values in the wrong format, making parameter repair the more productive edit at this tier. At the three-server level (Table~\ref{tab:subscore_3server}), parameter accuracy remains the largest gain for SmolLM3-3B ($+2.35$, from 1.33 to 3.68) and Llama-3.2-3B ($+1.30$, from 0.6 to 1.9). Dependency awareness, however, shows the smallest improvement at every tier: $+0.81$ at one server, $+0.59$ at two servers, and $+0.50$ at three servers for Llama-3.2-3B, reaching only 1.12 at the three-server level. SmolLM3-3B follows the same pattern, with dependency awareness reaching only 1.14 at three servers ($+0.65$). Typed edits are most effective at correcting tool selection and parameter errors; maintaining correct dependency chains across multiple servers remains a harder problem that local graph mutations address only partially.

\section{Component Ablation Studies}
\label{appendix:ablation}

We measure how much each EvoFlux component contributes to the observed improvements on the held-out validation split (43 tasks).

\subsection{Component Attribution}
\label{appendix:component_attribution}

EvoFlux combines three phases: (1)~initialization with error-guided retry (up to $R{=}10$ attempts, equivalent to Reflexion~\citep{shinnReflexionLanguageAgents2023}), (2)~evolutionary typed mutations that modify the population through execution-conditioned edits, and (3)~meta guidance that redesigns the workflow when progress stalls. We measure the incremental contribution of each phase by recording the best score achievable using only candidates from progressively larger phase subsets.

\begin{table}[t]
\centering
\footnotesize
\renewcommand{\arraystretch}{1.1}
\caption{Component ablation on the held-out split (43~tasks). Init+Retry uses up to $R{=}10$ error-guided retries ($\approx$~Reflexion). $\Delta$ columns show the incremental gain from each subsequent component. ``Best from Evo'' reports the fraction of queries whose best score was first achieved during the evolve or meta phase.}
\label{tab:ablation}
\resizebox{\columnwidth}{!}{%
\begin{tabular}{@{}l ccc cc c@{}}
\toprule
Model & \makecell{Init+\\Retry} & +Evolve & Full & $\Delta$Evo & $\Delta$Meta & \makecell{Best from\\Evo} \\
\midrule
DeepSeek-1.5B & 0.62 & 1.40 & 1.40 & +0.79 & +0.00 & 88\% \\
Llama-3.2-3B  & 0.87 & 2.29 & 2.45 & +1.42 & +0.16 & 93\% \\
SmolLM3-3B    & 1.13 & 2.09 & 2.15 & +0.96 & +0.07 & 88\% \\
Qwen3.5-4B   & 1.55 & 3.94 & 3.94 & +2.38 & +0.00 & 98\% \\
\bottomrule
\end{tabular}%
}
\end{table}

Table~\ref{tab:ablation} shows two things. The retry baseline (Init+Retry) already uses the full $R{=}10$ error-guided retry budget, making it a strong error-feedback baseline equivalent to Reflexion. Evolutionary mutations still contribute an additional +0.79 to +2.38 points on top of retry, accounting for 45--60\% of the final score depending on the planner. The contribution is lowest for SmolLM3-3B (45\%), where Init+Retry already captures a larger share, and highest for Qwen3.5-4B (60\%). Across all four planners, 88--98\% of queries achieve their best score during the evolve or meta phase, not during initialization.

\begin{figure}[t]
    \centering
    \includegraphics[width=\columnwidth]{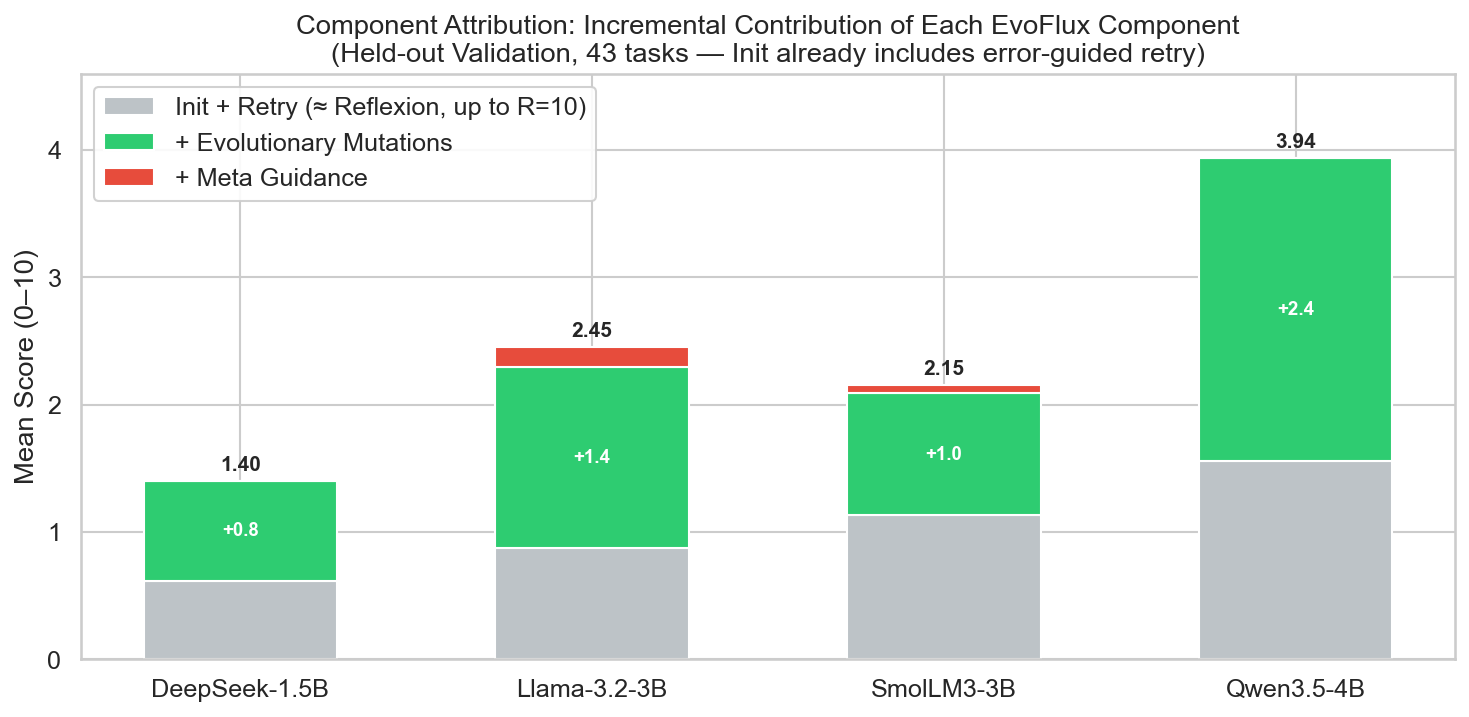}
    \caption{Stacked contribution of each EvoFlux component on the held-out split. Gray: Init+Retry ($\approx$~Reflexion). Green: evolutionary mutations. Red: meta guidance.}
    \label{fig:ablation_component}
\end{figure}

Figure~\ref{fig:ablation_component} visualizes these contributions as stacked bars. Meta guidance adds +0.07 for SmolLM3-3B and +0.16 for Llama-3.2-3B, and contributes nothing ($\Delta = 0.00$) for DeepSeek-1.5B and Qwen3.5-4B. It activates only when progress stalls and fires rarely enough that two of four planners never benefit from it in these runs. Evolutionary mutations drive the bulk of the improvement; meta guidance is a sparse escape hatch that helps only when the population is stuck. The attribution is correlational: phase credit records when the best score was first observed, so it reflects timing rather than causation.

\subsection{Exploration versus Exploitation}
\label{appendix:explore_exploit}

EvoFlux alternates between two mutation paths: \emph{exploration} (random typed edits, no LLM call) and \emph{exploitation} (LLM-guided edits conditioned on error evidence). Table~\ref{tab:explore_exploit} reports the mean candidate score and parent-improvement rate for each path.

\begin{table}[t]
\centering
\footnotesize
\setlength{\tabcolsep}{3pt}
\renewcommand{\arraystretch}{1.1}
\caption{Exploration (random typed edits, zero LLM cost) versus exploitation (LLM-guided edits) during the evolve phase. Improvement rate: fraction of mutations that score above their parent.}
\label{tab:explore_exploit}
\begin{tabular}{@{}l cc cc@{}}
\toprule
 & \multicolumn{2}{c}{Mean Score} & \multicolumn{2}{c}{Impr.\ Rate (\%)} \\
\cmidrule(lr){2-3}\cmidrule(lr){4-5}
Model & Explore & Exploit & Explore & Exploit \\
\midrule
DeepSeek-1.5B & 0.76 & 0.77 & 32.9 & 36.2 \\
Llama-3.2-3B  & 1.17 & 1.25 & 33.4 & 34.9 \\
SmolLM3-3B    & 1.16 & 1.10 & 31.7 & 29.6 \\
Qwen3.5-4B   & 1.90 & 2.28 & 35.6 & 42.5 \\
\bottomrule
\end{tabular}
\end{table}

The results show that random typed edits perform surprisingly close to LLM-guided edits for three of four planners. DeepSeek-1.5B (0.76 vs.\ 0.77), Llama-3.2-3B (1.17 vs.\ 1.25), and SmolLM3-3B (1.16 vs.\ 1.10, with exploration slightly \emph{outperforming} exploitation) all show narrow gaps or reversed advantages. Only Qwen3.5-4B shows a clear exploitation advantage (2.28 vs.\ 1.90, improvement rate 42.5\% vs.\ 35.6\%).

This pattern suggests that the typed edit structure itself, rather than LLM guidance, is the larger contributor for weaker planners. When the planner cannot reliably diagnose failures from error evidence, random edits constrained to valid operations perform about as well as guided ones. LLM guidance pays off primarily when the planner is strong enough to propose targeted fixes. The adaptive intensity controller balances these paths automatically, increasing exploration probability when progress stalls and favoring exploitation when recent mutations have been productive (Figure~\ref{fig:adaptive_intensity}).

\begin{figure*}[t]
    \centering
    \includegraphics[width=\textwidth]{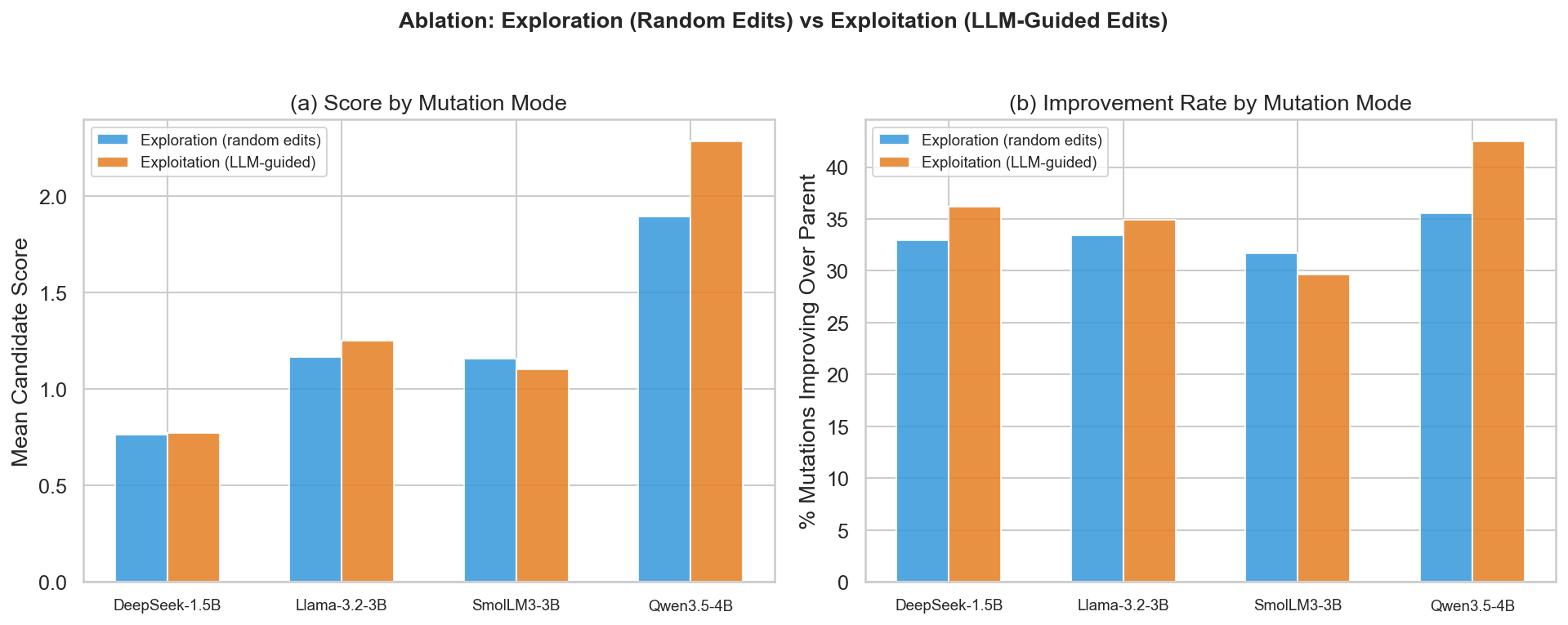}
    \caption{(a)~Mean candidate score and (b)~parent-improvement rate by mutation mode.}
    \label{fig:explore_exploit}
\end{figure*}

\subsection{Convergence Across Generations}
\label{appendix:convergence}

\begin{figure*}[t]
    \centering
    \includegraphics[width=\textwidth]{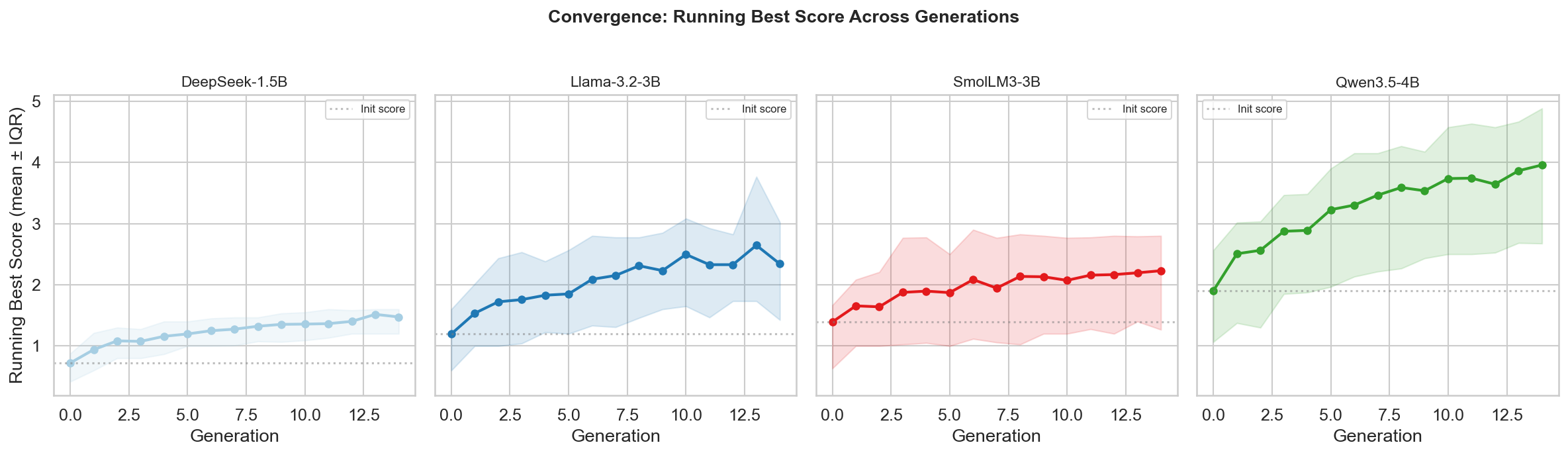}
    \caption{Running best score across generations (solid line: mean; shaded band: interquartile range over 43~tasks). The dashed line marks the init-phase score. Scores continue improving throughout the budget.}
    \label{fig:convergence}
\end{figure*}

Figure~\ref{fig:convergence} tracks the running best score across evolutionary generations. For Qwen3.5-4B, the running best rises from 1.90 at generation~0 to 3.96 at generation~14, with positive marginal gains continuing at late generations. (The generation index in Figure~\ref{fig:convergence} counts sequential candidate evaluations across all phases, so the x-axis represents cumulative search progress rather than phase boundaries. The Init+Retry scores in Table~\ref{tab:ablation} use phase labels, which count only initialization candidates. The convergence curve therefore tracks the running best score over all evaluations.) The same pattern holds across all four planners: the evolutionary loop adds value beyond early generations, and the absence of a sharp plateau suggests that additional budget could yield further improvements.

\begin{figure*}[t]
    \centering
    \includegraphics[width=\textwidth]{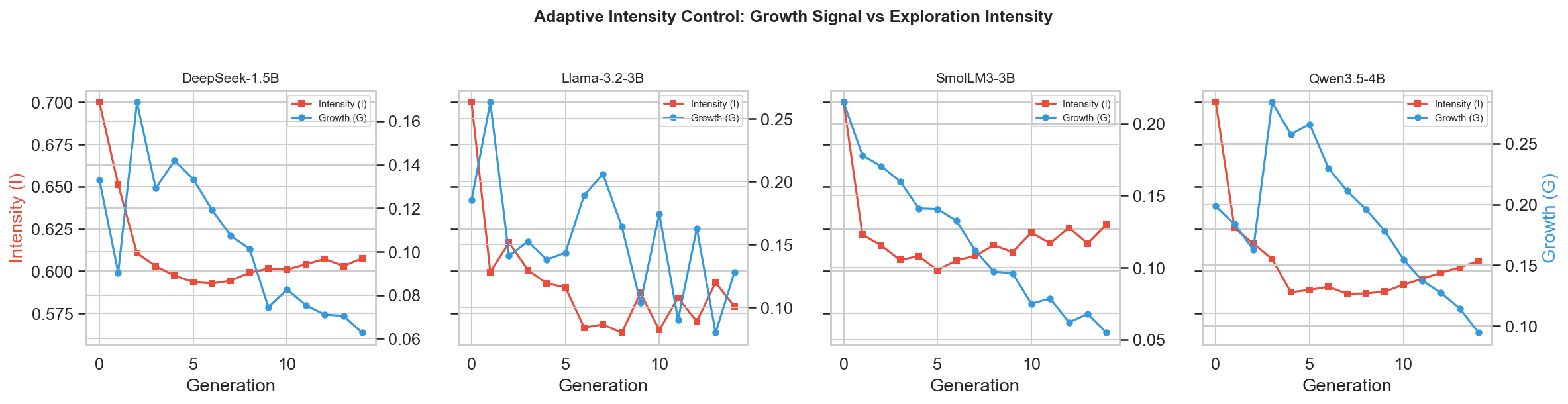}
    \caption{Adaptive intensity control across generations for each planner separately. Red line: exploration intensity~$I$. Blue line: growth signal~$G$. When progress stalls ($G$ low), intensity rises to explore more broadly.}
    \label{fig:adaptive_intensity}
\end{figure*}

Figure~\ref{fig:adaptive_intensity} shows the adaptive intensity controller in action. When the growth signal~$G$ drops (progress stalls), the intensity~$I$ rises, triggering more exploratory mutations. When $G$ increases (improvements found), $I$ drops, concentrating search on refining the current best. This feedback loop prevents premature convergence and avoids wasting LLM calls during exploration-heavy phases.

\section{Comparison with Simpler Search Strategies}
\label{appendix:baselines}

We compare EvoFlux against simpler search strategies to separate the contribution of evolutionary search from the benefit of generating multiple candidates.

\subsection{Best-of-N Sampling}
\label{appendix:best_of_n}

Best-of-N sampling generates $N$ independent workflow candidates and selects the highest-scoring one. We simulate this by drawing $N$ candidates uniformly at random from EvoFlux's own candidate pool (across all phases) and recording the maximum score, averaged over 50 trials per query.

Because the candidates are drawn from EvoFlux's pool rather than generated independently, this comparison tests whether evolutionary \emph{ordering} matters beyond simply having access to a diverse pool. A true best-of-N baseline would generate $N$ fresh plans from scratch, requiring $N$ independent planner calls per query and producing a different candidate distribution. The results below show that guided search over a fixed pool outperforms random selection from the same pool. They do not speak to how EvoFlux compares against independently generated best-of-N candidates.

\begin{figure*}[t]
    \centering
    \includegraphics[width=\textwidth]{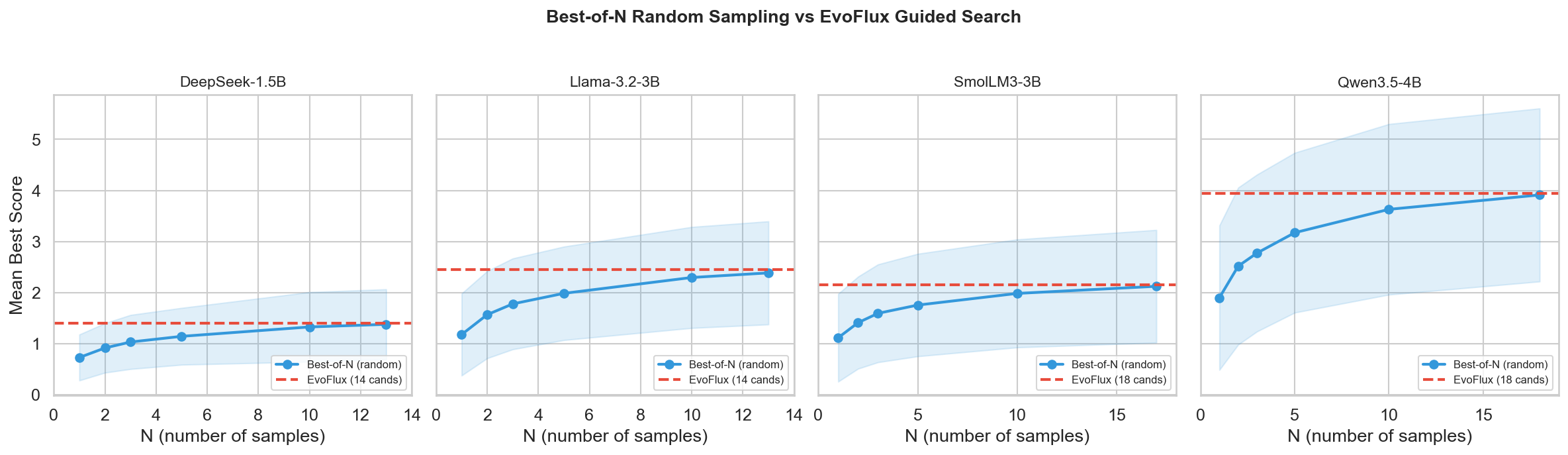}
    \caption{Best-of-N random sampling (blue curve $\pm 1\sigma$) vs.\ EvoFlux guided search (red dashed line) at matched candidate counts per query.}
    \label{fig:best_of_n}
\end{figure*}

Figure~\ref{fig:best_of_n} compares best-of-N (from the shared pool) against EvoFlux at matched candidate counts. EvoFlux outperforms random selection at the same $N$: for Qwen3.5-4B with 18 candidates per query, EvoFlux achieves 3.94 vs.\ 3.91 for best-of-18; for Llama-3.2-3B with 14~candidates, 2.45 vs.\ 2.39 for best-of-14. The gap is modest because EvoFlux's typed mutations produce correlated candidates (each builds on a parent), so random selection from the same pool already captures much of the diversity. The advantage of evolutionary search is that it directs mutations toward failing nodes rather than relying on random draws.

\subsection{Error-Guided Retry Baseline}
\label{appendix:retry_baseline}

The initialization phase of EvoFlux already implements error-guided retry: the planner generates a workflow, executes it, and if it fails, retries with the error context (up to $R{=}10$ attempts). This is equivalent to Reflexion~\citep{shinnReflexionLanguageAgents2023}. Table~\ref{tab:retry_vs_evoflux} directly compares this retry-only baseline against the full EvoFlux pipeline.

\begin{table}[t]
\centering
\footnotesize
\setlength{\tabcolsep}{3pt}
\renewcommand{\arraystretch}{1.1}
\caption{Error-guided retry ($\approx$~Reflexion, $R{=}10$) vs.\ full EvoFlux. Tok Ratio: how many more tokens EvoFlux uses relative to retry-only.}
\label{tab:retry_vs_evoflux}
\begin{tabular}{@{}l cc c cc@{}}
\toprule
Model & \makecell{Retry\\Score} & \makecell{EvoFlux\\Score} & Gain & \makecell{Tok\\Ratio} \\
\midrule
DeepSeek-1.5B & 0.62 & 1.40 & +0.79 & 3.0$\times$ \\
Llama-3.2-3B  & 0.87 & 2.45 & +1.58 & 3.2$\times$ \\
SmolLM3-3B    & 1.13 & 2.15 & +1.02 & 4.6$\times$ \\
Qwen3.5-4B   & 1.55 & 3.94 & +2.38 & 7.7$\times$ \\
\bottomrule
\end{tabular}
\end{table}

EvoFlux improves over retry by +0.79 to +2.38 points while using 3.0--7.7$\times$ more tokens. The extra compute goes to evolutionary mutations and meta guidance, not additional retry attempts. Population-based search with typed edits contributes beyond what error-guided retry alone provides.

\subsection{Compute-Matched Comparison}
\label{appendix:compute_matched}

\begin{figure}[t]
    \centering
    \includegraphics[width=\columnwidth]{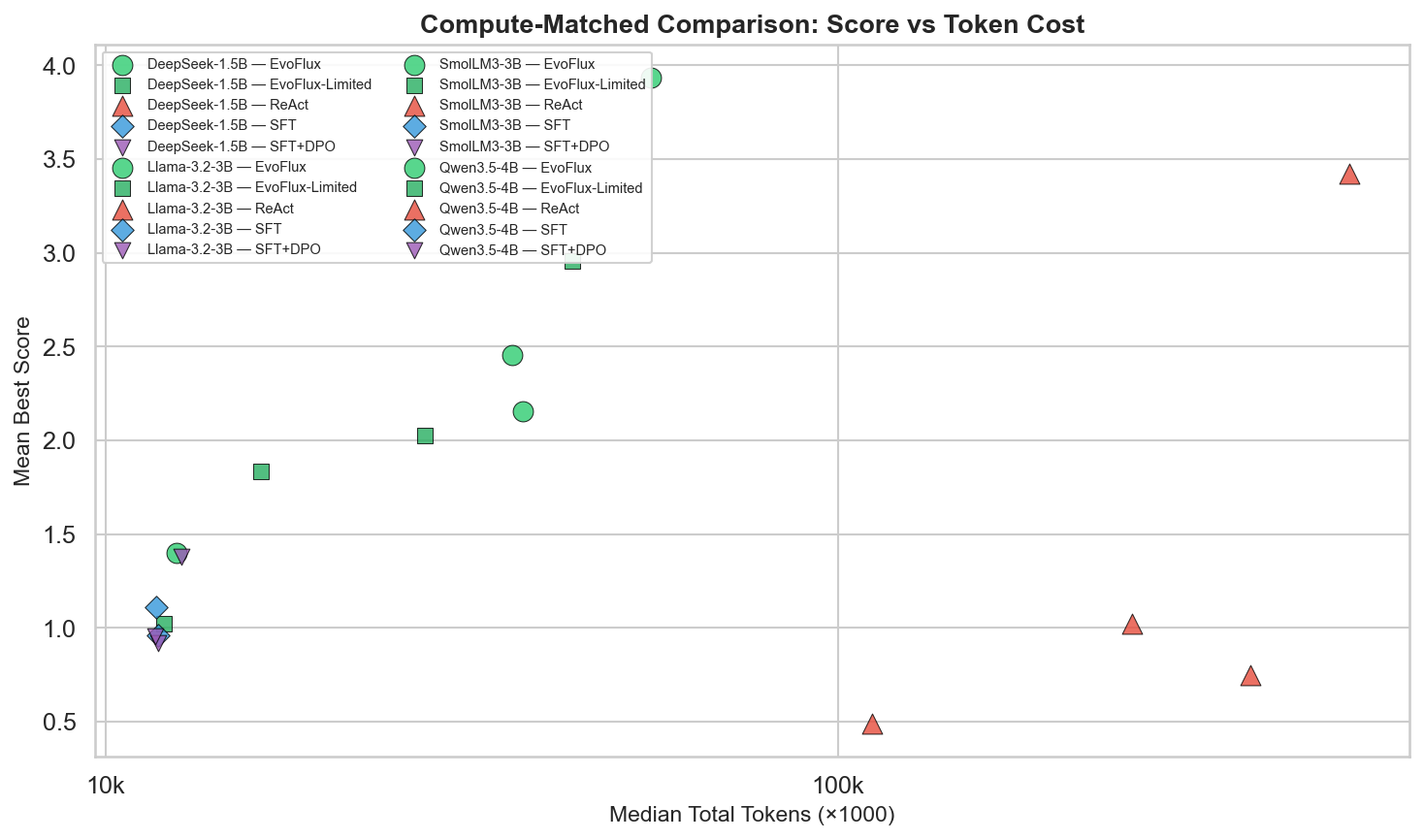}
    \caption{Score vs.\ median planner token cost across methods. EvoFlux achieves higher scores at lower token cost than ReAct on three planners. On Qwen3.5-4B, ReAct scores higher but costs $9\times$ more tokens.}
    \label{fig:compute_matched}
\end{figure}

We compare all methods at matched token costs to test whether EvoFlux provides an advantage at comparable inference budgets. Figure~\ref{fig:compute_matched} plots mean best score against median total planner tokens per query.

EvoFlux outperforms ReAct on three of four planners (DeepSeek-1.5B, Llama-3.2-3B, SmolLM3-3B). On Qwen3.5-4B, ReAct achieves a higher score but at $9\times$ the token cost (median planner tokens: 55k for EvoFlux vs.\ 500k for ReAct). We also evaluate EvoFlux under a reduced budget (``EvoFlux-Limited'', using the default hyperparameters without the expanded budget). Even at this reduced budget, EvoFlux-Limited outperforms ReAct on three of four planners while using $3$--$22\times$ fewer planner tokens.

\section{Graph Properties of the Workflows on the Validation Split}\label{appendix:depth_width}
\begin{figure*}[t]
    \centering
    \includegraphics[width=\textwidth]{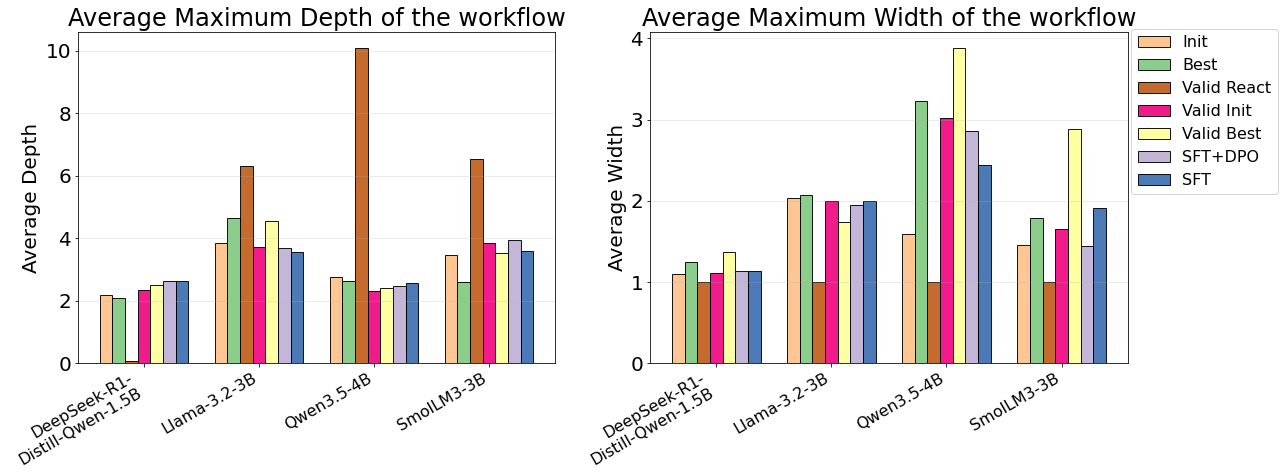}
    \caption{ Average workflow depth and width across stages.
    }
    \label{fig:depth_width}
\end{figure*}
Figure~\ref{fig:depth_width} compares the structural shape of generated workflows across the initial policy, EvoFlux, ReAct, and trained checkpoints. The left panel reports workflow depth, which reflects the length of sequential dependency chains. The right panel reports workflow width, which reflects the number of parallel or branch-like actions available at a workflow level. These two measurements separate chain complexity from breadth. Depth captures how far information must travel through ordered steps. Width captures how broadly the planner gathers, transforms, or branches over evidence.

ReAct often generates the deepest workflows. Its trajectory format appends each action to a growing sequence of prior thoughts, tool calls, and observations. Valid ReAct produces especially large depth for Qwen3.5-4B and remains among the deepest settings for the other planners. This extra depth should be interpreted carefully, since ReAct can produce strong outcomes with a capable planner while also increasing variance and token cost because each subsequent step accumulates more context.

EvoFlux shows a different profile. Best and Valid Best generally increase workflow width more than depth, especially on the validation side. Evolutionary repair broadens the workflow by adding or preserving evidence-gathering calls, lookup paths, and dependency-supporting steps instead of extending a long sequential chain. This matches the typed edit design, where search can insert missing evidence steps, remove harmful calls, redirect dependencies, and reorder execution without regenerating the entire graph.

The trained checkpoints show a weaker structural signal. SFT and SFT+DPO sometimes match or slightly exceed the initial workflow width, but that added structure does not yield reliable execution gains. Trained checkpoints can reproduce surface action patterns while failing to preserve executable dependency structure. Wider graphs become brittle when extra calls are not tied to valid intermediate outputs.

Workflow quality depends on how the graph structure carries evidence, not on graph size alone. ReAct builds deeper trajectories; EvoFlux broadens graphs through execution-guided repair. SFT and SFT+DPO can reproduce the visible shape of successful workflows without preserving the dependency logic that makes them executable.

\section{Workflow Structure Analysis}
\label{appendix:workflow_structure_analysis}

\begin{figure*}[t]
    \centering
    \includegraphics[width=\textwidth]{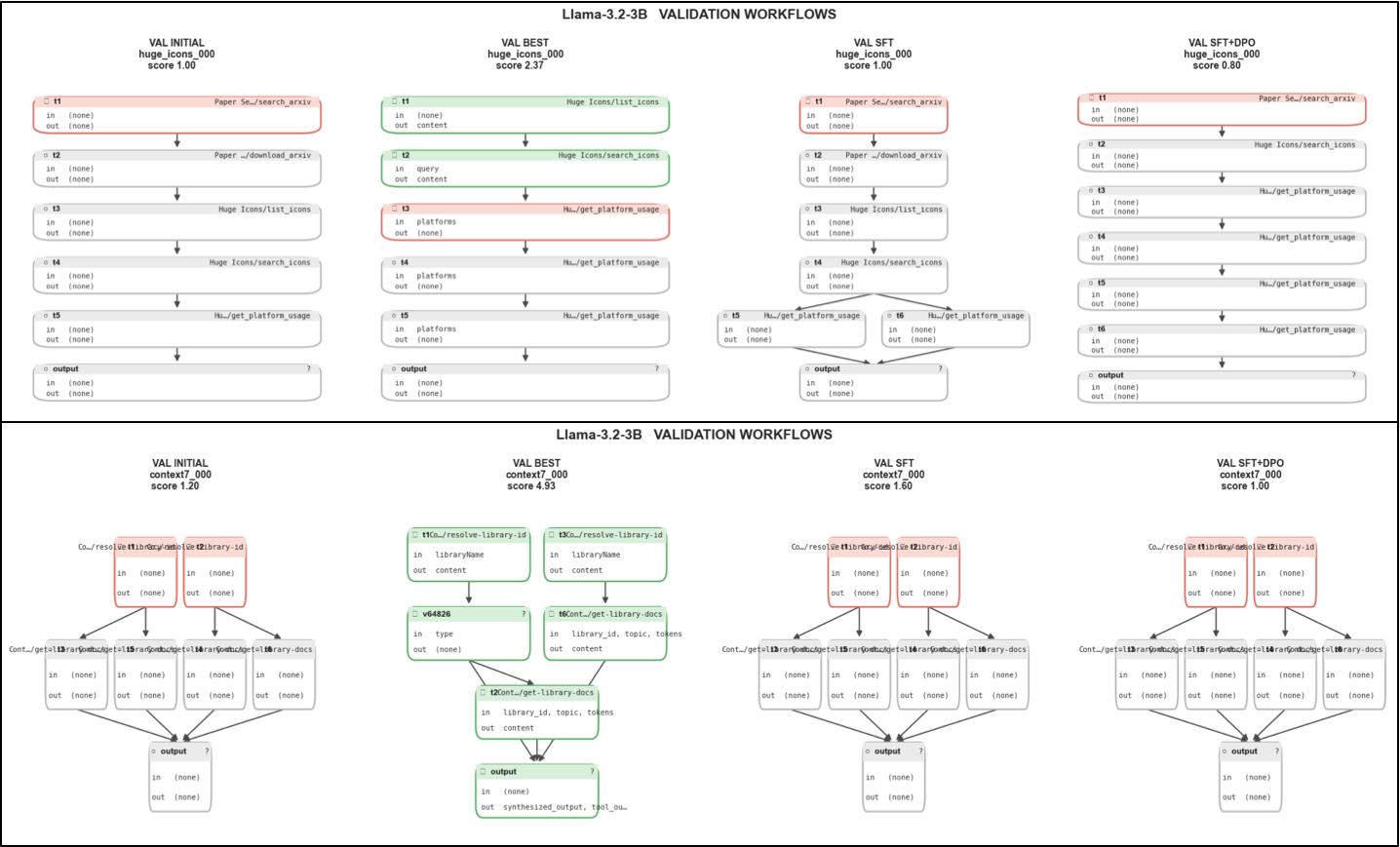}
    \caption{
    Representative validation workflows for Llama-3.2-3B under four regimes.
    EvoFlux produces stronger held-out workflows by selecting more relevant tools
    and repairing dependency flow, while SFT and SFT+DPO often imitate action patterns
    without preserving executable structure.
    }
    \label{fig_llama_validation_workflows}
\end{figure*}

Figure~\ref{fig_llama_validation_workflows} compares Llama-3.2-3B validation workflows across zero-shot inference, EvoFlux, SFT, and SFT+DPO. The examples illustrate the aggregate validation trend: Valid Best substantially outperforms the trained checkpoints. EvoFlux improves weak initial graphs through execution-guided repair, while SFT and SFT+DPO tend to preserve shallow tool-call templates without reliable workflow structure.

For \texttt{huge\_icons\_000}, the initial workflow scores 1.00 because it begins with irrelevant paper-search tools before reaching the Huge Icons server. EvoFlux raises the score to 2.37 by starting directly with icon listing and search, providing task-relevant evidence earlier. The remaining platform-usage calls still limit the final score, but the graph shows a clear repair in tool-family selection. By contrast, SFT repeats the paper-search pattern, and SFT+DPO expands into a longer chain of mostly unproductive calls, dropping to 0.80.

For \texttt{context7\_000}, the difference is bigger. The initial workflow scores 1.20 because it calls the right Context7 tools but weakly connects library resolution to documentation retrieval. EvoFlux reaches 4.93 by resolving library identifiers, feeding them into documentation calls, and routing retrieved content into synthesis before the final output. This graph succeeds because its edges carry usable evidence between steps. SFT and SFT+DPO keep the broad resolve-then-retrieve shape, but duplicate steps and weak dependencies keep their scores low.

The validation results support three conclusions. EvoFlux transfers beyond the search split: Valid Best recovers roughly the same score range as Best for \texttt{Llama-3.2-3B} and roughly two-thirds of the Best score for \texttt{Qwen3.5-4B}. Both Valid Best scores exceed the trained-checkpoint means by more than a factor of two.

Search-mined training data does not teach the small planners to reproduce the search process. SFT and SFT$+$DPO match or underperform Valid Init for both planners, and \texttt{Qwen3.5-4B} suffers complete planning collapse after training. The same execution histories that produce useful inference-time variants fail to encode reliable workflow search into the model weights.

The planner-token results show that inference-time evolution does not entail runaway costs. For \texttt{Llama-3.2-3B}, EvoFlux uses fewer planner tokens per query than ReAct on validation tasks. For \texttt{Qwen3.5-4B}, EvoFlux adds a modest planner-token overhead while delivering a much larger gain in score and feasibility. Compact workflow construction benefits more from execution-guided inference-time refinement than from supervised or preference training over the resulting traces.

\end{document}